\documentclass[sigconf, 9pt]{acmart}
\AtBeginDocument{%
  \providecommand\BibTeX{{%
    \normalfont B\kern-0.5em{\scshape i\kern-0.25em b}\kern-0.8em\TeX}}}

\copyrightyear{2023}
\acmYear{2023}
\setcopyright{rightsretained}
\acmConference[SenSys '23]{The 21st ACM Conference on Embedded Networked Sensor Systems}{November 12--17, 2023}{Istanbul, Turkiye}
\acmBooktitle{The 21st ACM Conference on Embedded Networked Sensor Systems (SenSys '23), November 12--17, 2023, Istanbul, Turkiye}\acmDOI{10.1145/3625687.3625811}
\acmISBN{979-8-4007-0414-7/23/11}

\usepackage{algorithm}
\usepackage{algorithmic}
\usepackage{amsthm}
\usepackage{amsfonts}
\usepackage{amsmath}
\usepackage{subcaption}
\usepackage{booktabs}
\usepackage{xspace}
\usepackage{multirow}
\usepackage{textcomp}
\usepackage{siunitx}
\usepackage{xcolor}
\usepackage{enumitem}
\usepackage[T1]{fontenc}

\newtheoremstyle{tight}
{2pt}
{2pt}
{}
{}
{\bfseries}
{\textit{:}}
{.5em}
{\textit{\thmname{#1}\thmnumber{ #2}}}

\theoremstyle{tight}
\newtheorem{corollary}{Corollary}

\begin{document}
\setlength{\floatsep}{0.6\baselineskip}
\setlength{\textfloatsep}{0.5\baselineskip}
\setlength{\dblfloatsep}{0.6\baselineskip}
\setlength{\dbltextfloatsep}{0.5\baselineskip}
\setlength{\belowcaptionskip}{0.5\baselineskip}
\setlength{\abovecaptionskip}{0.5\baselineskip}
\setlength{\abovedisplayskip}{2pt}
\setlength{\belowdisplayskip}{2pt}

\title{Physics-Informed Data Denoising for Real-Life Sensing Systems}

\author{Xiyuan Zhang}
\affiliation{%
  \institution{University of California, San Diego}
  \country{}
}
\email{xiyuanzh@ucsd.edu}

\author{Xiaohan Fu}
\affiliation{%
  \institution{University of California, San Diego}
  \country{}
}
\email{xhfu@ucsd.edu}

\author{Diyan Teng}
\affiliation{%
  \institution{Qualcomm}
  \country{}
}
\email{diyateng@qti.qualcomm.com}

\author{Chengyu Dong}
\affiliation{%
  \institution{University of California, San Diego}
  \country{}
}
\email{cdong@ucsd.edu}

\author{Keerthivasan Vijayakumar}
\affiliation{%
  \institution{University of California, San Diego}
  \country{}
}
\email{kevijayakumar@ucsd.edu}

\author{Jiayun Zhang}
\affiliation{%
  \institution{University of California, San Diego}
  \country{}
}
\email{jiz069@ucsd.edu}

\author{Ranak Roy Chowdhury}
\affiliation{%
  \institution{University of California, San Diego}
  \country{}
}
\email{rrchowdh@ucsd.edu}

\author{Junsheng Han}
\affiliation{%
  \institution{Qualcomm}
  \country{}
}
\email{junsheng@qti.qualcomm.com}

\author{Dezhi Hong}
\authornote{Work unrelated to Amazon.}
\affiliation{%
  \institution{Amazon}
  \country{}
}
\email{hondezhi@amazon.com}

\author{Rashmi Kulkarni}
\affiliation{%
  \institution{Qualcomm}
  \country{}
}
\email{rashmik@qti.qualcomm.com}

\author{Jingbo Shang}
\affiliation{%
  \institution{University of California, San Diego}
  \country{}
}
\email{jshang@ucsd.edu}

\author{Rajesh K. Gupta}
\affiliation{%
  \institution{University of California, San Diego}
  \country{}
}
\email{rgupta@ucsd.edu}

\newcommand{\our}{\mbox{PILOT}\xspace}
\newcommand{\smallsection}[1]{\vspace{1mm}\noindent\textbf{#1.}}
\newcommand{\CO}{CO$_2$~}

\renewcommand{\shortauthors}{Xiyuan Zhang et al.}

\begin{abstract}
Sensors measuring real-life physical processes are ubiquitous in today's interconnected world. 
These sensors inherently bear noise that often adversely affects performance and reliability of the systems they support. 
Classic filtering-based approaches introduce strong assumptions on the time or frequency characteristics of sensory measurements, 
while learning-based denoising approaches typically rely on using ground truth clean data to train a denoising model, which is often challenging or prohibitive to obtain for many real-world applications.
We observe that in many scenarios, the relationships between different sensor measurements (e.g., location and acceleration) are analytically described by laws of physics (e.g., second-order differential equation).
By incorporating such physics constraints, we can guide the denoising process to improve even in the absence of ground truth data.  
In light of this, we design a physics-informed denoising model that leverages the inherent algebraic relationships between different measurements governed by the underlying physics. By obviating the need for ground truth clean data, our method offers a practical denoising solution for real-world applications. We conducted experiments in various domains, including inertial navigation, CO$_2$ monitoring, and HVAC control, and achieved state-of-the-art performance compared with existing denoising methods. 
Our method can denoise data in real time (4ms for a sequence of 1s) for low-cost noisy sensors and produces results that closely align with those from high-precision, high-cost alternatives, leading to an efficient, cost-effective approach for more accurate sensor-based systems. 

\end{abstract}



\maketitle

\begin{figure}
    \centering
    \includegraphics[width=0.62\linewidth]{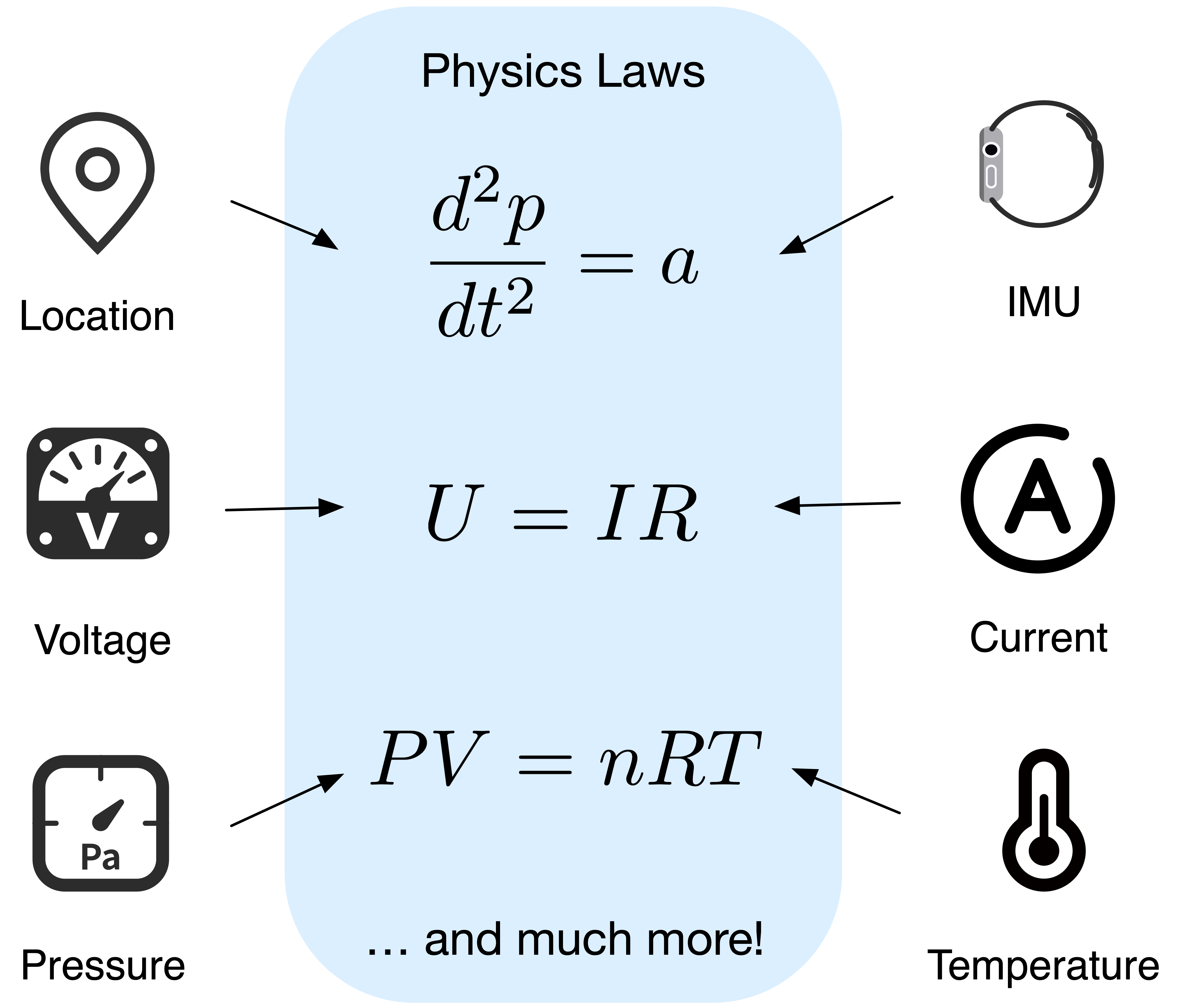}
    \caption{Real-life data captured by sensors are often governed by the laws of physics.}
    \label{fig:intro}
\end{figure}

\section{Introduction}

Sensors measuring various real-life physical processes have permeated our daily lives. These sensors play a crucial role in acquiring a large amount of data in various applications including environmental monitoring, healthcare, smart home and building, and transportation, enabling context inference, pattern recognition, and informed downstream decision-making.
However, because of factors such as environmental interference, electrical fluctuations, and imprecision of the sensor itself, sensor data are naturally noisy. Such noise degrades data quality and adversely affects performance of downstream applications.

To improve sensor data quality, there has been a line of research on noise reduction, from traditional filtering approaches that rely on prior knowledge of signal characteristics in time or frequency domain~\cite{de2017insights,omitaomu2011empirical}, to more advanced machine learning methods~\cite{wang2021tstnn,zhang2017beyond}. Existing machine learning-based methods typically assume availability of ground truth clean data in order to train a denoising model in a supervised manner. However, given the inherently noisy nature of sensors, we often do not have access to ground truth \emph{clean} data in real-world applications. In addition to supervised denoising methods, researchers have also developed self-supervised denoising methods, mostly in computer vision applications~\cite{lehtinen2018noise2noise,moran2020noisier2noise}. These approaches often make simplified assumptions (e.g., about noise distributions) that may not accurately reflect real-world sensor data with complex correlations. Moreover, they do not leverage the unique physical characteristics of sensor data distinct from typical image or text data. 

\begin{figure*}[t]
    \centering
    \includegraphics[width=0.98\linewidth]{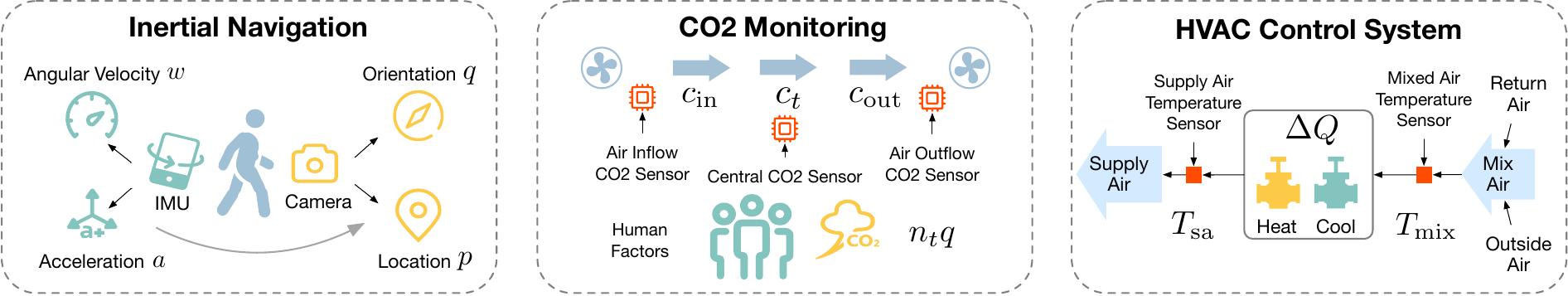}
    \caption{Three example applications of \our. Inertial Navigation (left): \our denoises inertial measurements (angular velocity and acceleration) and camera measurements (location and orientation) based on the motion law; CO$_2$ Monitoring (mid): \our denoises CO$_2$ readings leveraging the CO$_2$ relationship in a room; HVAC Control System (right): \our denoises the supply and mix air temperature sensors of an AHU utilizing the heat transfer equation.}
    \label{fig:prelim}
\end{figure*}

In multiple applications, we observe that different measurement channels of sensor data collected from real-life physical processes are often correlated and characterized by equations rooted in well-understood underlying physics. We illustrate three example scenarios in Figure~\ref{fig:intro}. In the first example, the relationship between location and acceleration is captured by the motion law, which equates the second-order derivative of location to acceleration. The second example connects the measurements by drawing upon Ohm's law -- the ratio of voltage and current is a constant, resistance, of the circuit. In the third example, the ideal gas law establishes a link between pressure and temperature measurements. These three examples are merely a sample of the numerous physics principles that govern real-life sensing systems. This paper builds upon the intuition that \emph{such physics-based principles among measured channels can be used as constraints to improve sensor denoising techniques}.

\begin{figure}
    \centering
    \includegraphics[width=0.98\linewidth]{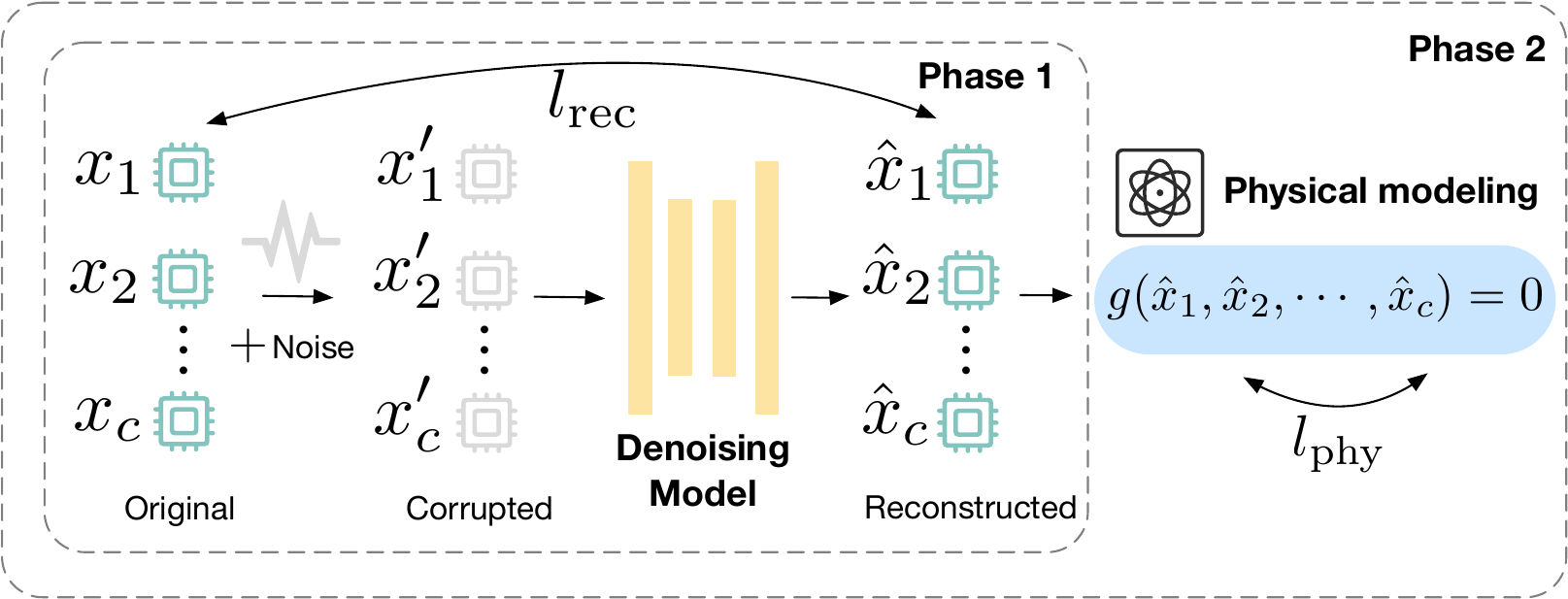}
    \caption{Physics-Informed denoising model framework. During training, we manually add noise to the original sensor data, and minimize a reconstruction loss to remove such noise. We also integrate a physics-based loss to ensure that reconstructed data obey the law of physics.}
    \label{fig:model}
\end{figure}

To this end, we propose a physics-informed denoising method, \our (Physics-Informed Learning for denOising Technology), by incorporating analytical model based on known physics as loss functions during training. Specifically, we introduce physics-derived equations as soft constraints that the denoised output from our model needs to satisfy. 
We illustrate the model architecture in Figure~\ref{fig:model}. 
We intentionally add noise to corrupt the original data collected from multiple sensors. Feeding this corrupted data as input, we train a denoising model to remove the added noise by minimizing a \textit{reconstruction loss} between the model output and the original undistorted input. However, since the original sensor data are inherently noisy, a reconstruction loss alone is insufficient to remove such noise. Therefore, we introduce an additional \textit{physics-based loss} to minimize physics misalignment of denoised output. For the example of inertial navigation, the corresponding physics equations include differential equations between position/orientation and acceleration/angular velocity. 
Therefore, in this example, the physics constraint would penalize discrepancy between the derivatives of location/orientation and the acceleration/angular velocity. Denoising processes for other applications follow similarly via a unified mathematical formulation that we will elaborate on.

To evaluate our approach, we conducted experiments on different real-world scenarios including inertial navigation, CO$_2$ monitoring, and HVAC (Heating, Ventilation, and Air Conditioning) control. 
For inertial navigation, \our generates denoised results that are most coherent to physics equations, leading to the best performance in downstream inertial navigation applications on the benchmark OxIOD dataset~\cite{chen2018oxiod}. We collected data for the other two applications, detailed in Section~\ref{sec:co2} and Section~\ref{sec:hvac}. For CO$_2$ monitoring, we have two types of CO$_2$ sensors--one is low-cost and very noisy, and the other is more accurate but much more expensive. Denoised CO$_2$ measurements from the low-cost sensors by \our closely match those from the much more expensive CO$_2$ sensors. Consequently, our denoising technique offers significant cost savings. We observed similar advantage for HVAC control system, where our denoised temperature data collected from low-cost noisy sensors best match those collected from industry-grade sensors. 
In summary, this paper makes the following contributions: 
\begin{itemize}[nosep,leftmargin=*]
    \item We propose a novel denoising method, \our, trained under the guidance of physics constraints. To our best knowledge, this is the first \emph{generic}, physics-informed denoising method that supports different sensing applications.
    \item \our offers a \emph{practical} denoising solution for application scenarios where obtaining ground truth clean data or understanding the underlying noise distribution is challenging, better representing real-world sensing environments.
    \item We collected two datasets on CO$_2$ monitoring and HVAC control, with pairs of noisy and accurate sensor data.
    \item Extensive experiments in three real-world applications demonstrate that \our produces results that are closest to the clean data and most coherent to the laws of physics, achieving state-of-the-art denoising performance. We also deployed the model on edge devices and showed real-time denoising capability.
\end{itemize}

\section{Related Work}

\subsection{Physics-Informed Machine Learning}

\subsubsection{General Physics-Informed Machine Learning} The concept of Physics Priors refers to our understanding of the inherent principles of the physical world, taking various representations such as differential equations, symmetry constraints, and common-sense knowledge, among others. An emerging field, Physics-Informed Machine Learning (PIML), aims to incorporate physics priors into machine learning models~\cite{hao2022physics,wang2021physics,karniadakis2021physics}. A line of research has been proposed for such integration, which can be categorized into different directions: incorporating physics priors as loss regularization~\cite{raissi2019physics,ren2018learning,gao2021super,kelshaw2022physics}, imposing strict constraints into the model architectures~\cite{wang2020towards,takeishi2021physics,yang2022learning} or hybrid approaches~\cite{garcia2019combining}. In general, physics-informed machine learning methods enhance efficacy, generalizability, and robustness of the models. However, most works in the PIML domain evaluate on simplified synthetic datasets. 

\subsubsection{Physics-Informed Machine Learning for Sensing Systems} 
Several works (such as PIP~\cite{yi2022physical}, PACMAN~\cite{nagarathinam2022pacman}, Reducio~\cite{cao2022reducio}, and PhyAug~\cite{luo2021phyaug}) employ physics in practical settings. 
However, most approaches are domain-specific~\cite{kim2021physics,chen2018pga,falcao2021piwims,he2020scsv2}, and miss the consideration of physical relationships among different sensor measurement channels that can assist with the denoising process.
Multimodal learning methods~\cite{jeyakumar2019sensehar,lu2020milliego,wang2022capricorn,zhang2023modeling} are also relevant topics as different modalities mutually compensate and offer enriched information, but existing multimodal methods mostly focus on fusing multiple modalities for downstream prediction tasks (e.g., human activity recognition) without addressing the denoising challenge. 
To address this gap, we present the first physics-informed machine learning model that incorporates these relationships into denoising sensor data, pushing the capability of what is currently achievable in PIML.

\begin{figure*}
      \begin{subfigure}[b]{0.56\textwidth}
          \centering
          \includegraphics[width=\textwidth]{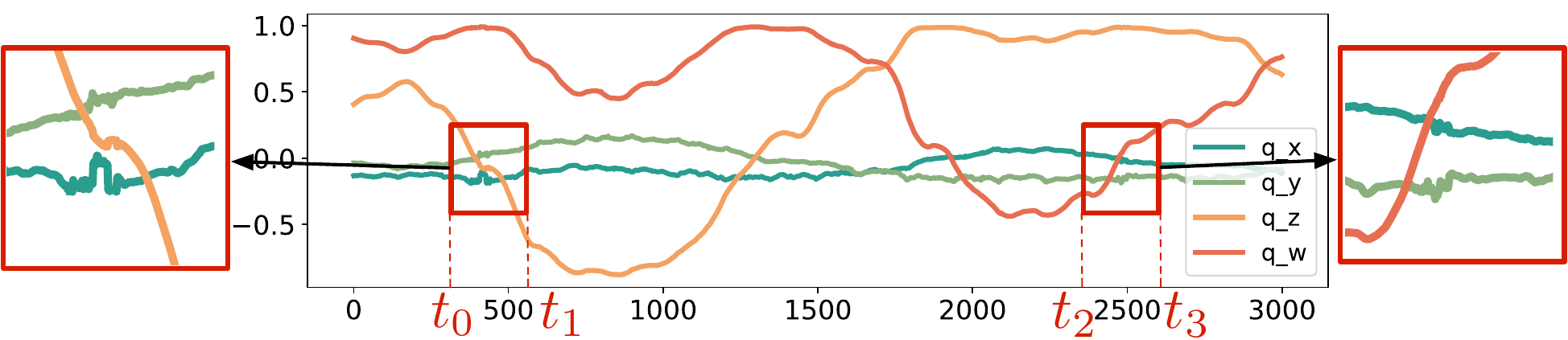}
          \caption{Noisy Data}
          \label{fig:prelim-quat}
      \end{subfigure}
      \centering
      \begin{subfigure}[b]{0.4\textwidth}
          \centering
          \includegraphics[width=\textwidth]{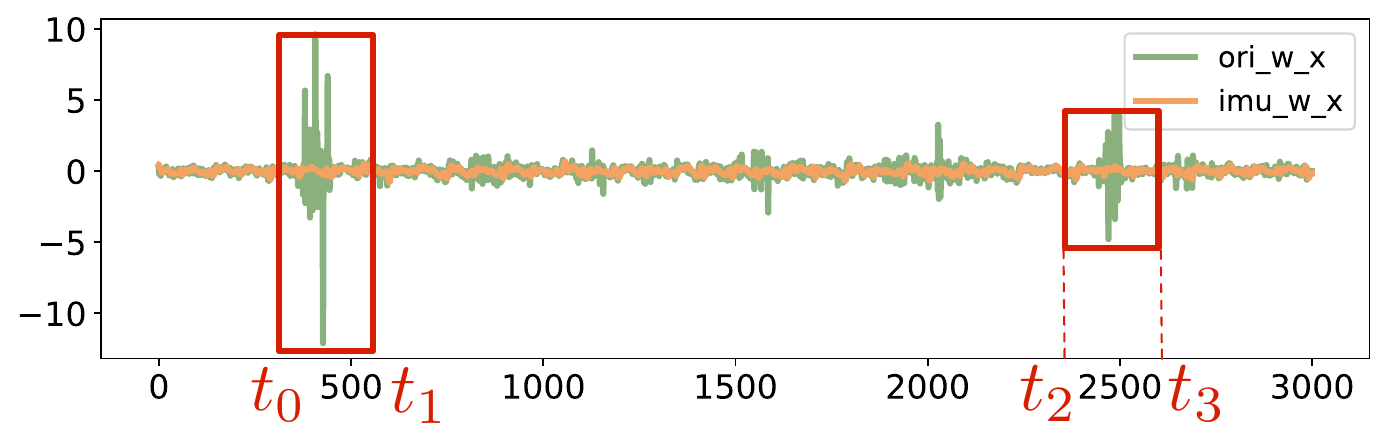}
          \caption{Measured  vs Computed Angular Velocity}
          \label{fig:prelim-imu}
      \end{subfigure}
         \caption{(a) Original orientation measurements. (b) Comparison of gyroscope measurements (imu\_w\_x) and angular velocities derived from orientations (ori\_w\_x) in x-axis. Regions with higher noise level ($t_0 \sim t_1, t_2 \sim t_3$) in the orientation data map to regions with large misalignment between IMU measurements and derived angular velocities. X-axis: timestep, Y-axis: meter.} 
\end{figure*}

\subsection{Denoising Methods}

Sensor data are often noisy due to intrinsic characteristics such as thermal noise, operating point drifts, or environmental effects. Existing works have proposed methods to denoise sensor data or adapt the model to noisy measurements~\cite{faulkner2013fresh,li2022sqee}. The majority of these strategies, however, is custom-tailored to specific domains of application, such as COTS WiFi-based motion sensing~\cite{zhu2017calibrating}, image data processing under various weather conditions~\cite{chen2020adaptive}, speech recognition~\cite{liu2021wavoice}, and clinical data analysis~\cite{park2017glasses}. Some researchers have also proposed more general denoising approaches through traditional estimation methods, or time-frequency domain analysis such as discrete wavelet transform~\cite{de2017insights} and empirical mode decomposition~\cite{omitaomu2011empirical}. However, traditional estimation methods such as Kalman smoother rely heavily on the assumption of Gaussian noise and Markovian transition, but real-world physics models such as human motion typically have a higher-order temporal correlation. These methods also rely on oversimplified assumptions that observed variables are conditionally independent over time. More recent deep learning approaches~\cite{zhang2017beyond,wang2021tstnn} are supervised methods that require ground truth clean data, and they neglect the unique benefits provided by physics-rich real-life sensing systems. Self-supervised denoising methods have been recently proposed to train denoising model without access to ground truth clean data~\cite{ulyanov2018deep,lehtinen2018noise2noise,huang2021neighbor2neighbor,moran2020noisier2noise,xu2020noisy}. These methods are mainly designed for the computer vision domain and make assumptions about the noise.

\subsection{Real-Life Sensing Systems}
There exist numerous physics-rich real-life sensing applications. In this study, we focus on three applications: 
inertial navigation, CO$_2$ monitoring, and HVAC systems. Inertial navigation predicts location and orientation based on IMU data, which is prone to erroneous pose estimate rapidly over time~\cite{shen2018closing}.
Classic approaches incorporate external measurements such as GNSS and camera, to periodically correct the propagated pose value, at the cost of additional hardware and power. Recently, learning-aided approaches become popular~\cite{shen2016smartwatch,liu2019real,yan2018ridi,chen2018ionet,herath2020ronin,liu2020tlio,sun2021idol,liu2022real}. CO$_2$ monitoring monitors \CO level to ensure occupant comfort. Researchers have designed data collection~\cite{sun2022c} and generation~\cite{weber2020detecting} methods for \CO data. When modeling \CO data, existing works have developed data-driven, differential equation models~\cite{weekly2015modeling}. Modeling \CO levels is beneficial for occupancy prediction and energy savings~\cite{arief2017hoc,10.1145/3408308.3431113}. HVAC Control System controls heating, ventilation and air conditioning based on temperature and airflow measurements~\cite{beltran2014optimal,nagarathinam2015centralized}. 
Apart from traditional model-based methods, recently deep reinforcement learning has also shown promising performance in modeling HVAC systems~\cite{kurte2021comparative,zhang2019building,xu2022accelerate}. 
\section{Preliminary}
\subsection{Problem Formulation}\label{sec:def}
\noindent \textbf{Clean Data.} We denote the unobserved ground truth clean dataset as $\mathcal{D}_X=\{\mathbf{X}_i\}_{i=0}^N$ which consists of $N$ samples. Each sample $\mathbf{X}_i \sim \mathcal{X}, \mathbf{X}_i \in \mathbb{R}^{c \times T}$,  where $\mathcal{X}$ denotes the clean data space, $T$ denotes the number of timesteps in the sensor data, and $c$ denotes the number of sensor measurement channels. Each sample $\mathbf{X}_i = [\mathbf{x}_{i1},\mathbf{x}_{i2},\cdots,\mathbf{x}_{ic}]^\intercal$, where $\mathbf{x}_{ij} \in \mathbb{R}^T$ represents the $j$th sensor measurement channel for the $i$th sample in the ground truth clean dataset. 
\smallskip

\noindent \textbf{Noisy Data.} We denote our observed noisy sensor dataset as $\mathcal{D}_Y=\{\mathbf{Y}_i\}_{i=0}^N$, composed of $N$ samples. Each sample $\mathbf{Y}_i \sim \mathcal{Y}, \mathbf{Y}_i \in \mathbb{R}^{c \times T}$, and  $\mathbf{Y}_i = [\mathbf{y}_{i1},\mathbf{y}_{i2},\cdots,\mathbf{y}_{ic}]^\intercal$, where $\mathcal{Y}$ denotes the noisy data space, $\mathbf{y}_{ij} \in \mathbb{R}^T$ represents the $j$th sensor measurement channel for the $i$th sample in the noisy dataset. \smallskip

\noindent \textbf{Noise.} Each noisy sample $\mathbf{Y}_i$ is corrupted from clean sample $\mathbf{X}_i$ by noise $\mathbf{\epsilon}_i \sim \mathcal{E}, \mathbf{\epsilon}_i \in \mathbb{R}^{c \times T}$, where $\mathcal{E}$ denotes the noise space. Specifically, $\mathbf{Y}_i = \mathbf{X}_i + \mathbf{\epsilon}_i, i=1,2,\cdots,N.$ \smallskip

\noindent \textbf{Physics Equations.} Equation $g(\cdot)$ describes the relationships between different sensor measurement channels in the ground truth clean data, i.e., $g(\mathbf{x}_{i1},\mathbf{x}_{i2},\cdots,\mathbf{x}_{ic})=0, i=1,2,\cdots,N.$  \smallskip

\noindent \textbf{Denoising Model.} The denoising model is parameterized by $\theta$ and maps noisy data to clean data by mapping $f(\cdot;\theta): \mathcal{Y} \rightarrow \mathcal{X} \subset \mathbb{R}^{c \times T}.$ We incorporate the physics equation $g(\cdot)$ as a constraint to optimize $f(\cdot;\theta)$ during training. 

\subsection{Physics Principles in Sensing Systems}
In this section, we introduce the underlying physics principles of three example sensing applications: inertial navigation, CO$_2$ monitoring, and HVAC control. 
\our leverages these physics equations as constraints during training to better guide the denoising process. We evaluate the denoising performance of these three sensing applications in Section~\ref{sec:eval}.
The major difference across different applications is the specific form of the physics constraint. Consequently, to adapt our model to new applications, the only requirement is to formulate the equation representing the physics relationship between different sensors in the new application. Thereafter, we can seamlessly incorporate the constraint, treat it as regularization, and maintain a uniform approach across varying applications.

We select these three use cases to demonstrate the different capabilities of the framework to model various physics relationships. The use cases cover different sensor types, domains, and scales. In particular, we choose the three applications to validate that the proposed method is robust across different scales --- from single-human interactions to building air circulation models across many zones and users. Further, we seek to cover as many types of sensors (IMU, Camera, CO$_2$, airflow, temperature, etc.) as possible in these three experiments to demonstrate the potential to generalize to various types of sensors and noise levels. Our method can incorporate different physical laws with varying degrees of physics model complexity. We also picked the applications where physics models are informative but imperfect due to various modeling challenges. The selected applications are known for their dependencies on additional measurement modalities to compensate for the imperfect physics model, such as the visual information for navigation and IMU physics. This motivates the necessity for both learning component- as well as physics-based models. 

\subsubsection{Application Scenario I: Inertial Navigation}
Inertial navigation aims to estimate a moving device's position and orientation based on IMU sensor measurements, shown in Figure~\ref{fig:prelim} (left). The inertial navigation model takes IMU measurements as input, and predicts locations and orientations. The ground truth location and orientation data are captured by cameras. However, both IMU and camera data carry noise due to factors like jittering or occlusion, posing challenges to training accurate inertial navigation models. If we denote location, orientation (in quaternion form), angular velocity, and acceleration as $p, q, w, a$ respectively, then their relationships can be mathematically expressed by equations $g_1(\cdot), g_2(\cdot)$:
\begin{equation}\label{eq:accel}
    g_1(a,p,q) = a  - R_q^T (\frac{d^2p}{dt^2} - g_0),
\end{equation}
\begin{equation}\label{eq:quat}
    g_2(w,q) = \frac{dq}{dt} - \frac{1}{2} q \otimes w,
\end{equation}
where $R_q, g_0, \otimes$ represent rotation matrix, gravity constant and quaternion multiplication. As a motivating example, we randomly select 30-second sensor data from the OxIOD dataset~\cite{chen2018oxiod} and visualize the orientation measurements in Figure~\ref{fig:prelim-quat}. We note two major noise-heavy regions around the $500^{th}$  timestep and the $2500^{th}$ timestep  (indicated by boxed regions). Applying Equation~\ref{eq:quat}, we derive the angular velocity from the orientation data, and compare it with the angular velocity directly measured by IMU gyroscope in Figure~\ref{fig:prelim-imu}. The misalignment between these two angular velocities appears around 
the same timesteps of noisy regions in the original orientation data. This highlights the role of known physics to guide the learning process of denoising models and its use in our model.

\subsubsection{Application Scenario II: CO$_2$ Monitoring}

CO$_2$ monitoring is of wide interest in smart building applications. An accurate zone CO$_2$ reading can enable transmission risk analysis~\cite{Rudnick2003-uo}, occupancy estimation~\cite{amt-12-1441-2019}, air quality monitoring~\cite{co2_airquality}
for energy-saving or residence safety purposes~\cite{10.1145/3408308.3431113}. Figure~\ref{fig:prelim} (middle) depicts one example placement of three CO$_2$ sensors proximate to the air intake, outtake, and the center of a standard office room. Assume $c_t, c_0, c_{\mathrm{in}}^t, c_{\mathrm{out}}^t$ represent the current average CO$_2$ concentration in the room, initial average CO$_2$ concentration in the room, CO$_2$ concentration for the input airflow, CO$_2$ concentration for the output airflow, respectively. Furthermore, let $v, V, q, n_t$ represent the airflow velocity, room volume, CO$_2$ emission rate per individual, and the number of occupants in the room. Note that all these extra parameters are known information of the environment. They may change over time, but their values at each time step are known. Then, we can write their relationship in the following equation $g(\cdot)$: 
\begin{equation}\label{eq:co2}
\begin{split}
    &  g(c_t, c_{\mathrm{in}}^t, c_{\mathrm{out}}^t) = c_t V - (c_0 V + \sum_t (c_{\mathrm{in}}^t v \Delta t)  \\
    & + \sum_t (n_t q \Delta t) - \sum_t (c_{\mathrm{out}}^t v \Delta t))
\end{split}
\end{equation}

Unfortunately, perfect CO$_2$ monitoring would require deploying multiple high-resolution and low-noise lab-level CO$_2$ sensors, which are prohibitively expensive for a typical commercial building consisting of hundreds of zones. By contrast, low-cost CO$_2$ sensors are readily available on the market with more noise and less accuracy (e.g., CCS881 as we will detail in Sec~\ref{sec:co2}).  We aim to incorporate the aforementioned physics equation to denoise CO$_2$ data collected from these low-cost sensors, such that data quality after our denoising algorithm can match the expensive counterparts. 

\subsubsection{Application Scenario III: Air Handling in HVAC Control System}
HVAC control systems manage critical parameters such as temperature, humidity, and air quality of commercial or residential buildings to ensure thermal comfort and optimal air quality for occupants while maintaining energy efficiency. Air Handling Units (AHUs), the core of an HVAC control system, take in outside air and return air from zones in the building, recondition it via heating or cooling, de-humidification, and then supply it to spaces. Downstream equipment such as Variable Air Volumes (VAVs) may further condition the air based on factors including occupancy, time of day, and specific temperature desired of the space. 
Figure~\ref{fig:prelim} (right) illustrates an AHU in an example HVAC control system with both heating and cooling functionality. The AHU mixes a certain ratio of return air from the conditioned spaces and the fresh outside air, and reconditions the mixed air through cooling and heating coils and supply air to various spaces. We denote the enthalpy difference across the heating and cooling coils by $\Delta Q$. The air's mass undergoing reconditioning is denoted as $m$, its specific heat capacity as $c$, the temperature of the supply air and mix air as $T_{\mathrm{sa}}$ and $T_{\mathrm{mix}}$, respectively. We then derive the following heat transfer equation under the assumption of no additional heat source aside from the cooling and heating coils: 
\begin{equation}\label{eq:hvac}
    g(\Delta Q,m,c,T_{\mathrm{sa}},T_{\mathrm{mix}}) = \Delta Q - mc(T_{\mathrm{sa}} - T_{\mathrm{mix}})
\end{equation}
Note that air mass $m$ can be acquired through the airflow rate (measured in Cubic Feet per Minute, or CFM) going across the AHU. Furthermore, the air's specific heat capacity, $c$, is both humidity and temperature dependent.

In practice, both supply air temperature and mix air temperature measurements are subject to noise, which impedes accurate control in the HVAC system. Therefore, we incorporate the above equation to denoise the two temperature measurements, enhancing the accuracy and reliability of the HVAC control system. 

\section{Methodology}

\subsection{Model Framework}
\noindent \textbf{Noise Injection.} We first follow the setting of typical supervised denoising methods to manually inject noise and train the denoising model to remove the injected noise. Given the unknown nature of real-world noise distributions, we first randomly sample noise from pre-defined distributions as simulations. This process is insufficient to capture the complex noise distributions, so we augment the model training with physics modeling process illustrated later. When sampling the noise, we use Gaussian distribution with zero mean and variance of $\sigma^2$ as an example and compare with other noise distributions in Section~\ref{sec:sensitivity} (Sensitivity Analysis). Mathematically, for each sample $\mathbf{Y}_{i}$ in our noisy dataset, we further sample Gaussian noise $\mathbf{n}_{i}$ and add it upon $\mathbf{Y}_{i}$ to obtain data $\mathbf{Z}_{i}$ with a higher degree of noise, i.e., $\mathbf{Z}_{i} = \mathbf{Y}_{i} + \mathbf{n}_{i}, \mathbf{n}_{i} \sim \mathcal{N}(0,\sigma^2).$ Note that our physics modeling component does not pose any constraints or assumptions on the noise type. The inclusion of Gaussian additive noise is purely for simulation purposes to facilitate the training of the denoising autoencoder. \smallskip

\noindent \textbf{Naive Denoising Model.} The naive denoising model performs optimization without modeling physical relationships. The denoising model takes data with manual noise injected $\mathbf{Z}_{i}$ as input, and the training objective is to output data with such manual noise removed, by minimizing the reconstruction loss, i.e.,
\begin{gather}
    \mathrm{argmin}_{\theta} \mathbb{E}_{\mathbf{Y}_{i} \sim \mathcal{Y}, \mathbf{n}_{i} \sim \mathcal{N}}\ell_{\mathrm{rec}}(f(\mathbf{Y}_{i} + \mathbf{n}_{i};\theta),\mathbf{Y}_{i}), \label{eq:goal} \\
    \ell_{\mathrm{rec}}(f(\mathbf{Y}_{i} + \mathbf{n}_{i};\theta),\mathbf{Y}_{i}) = \lvert\lvert f(\mathbf{Y}_{i} + \mathbf{n}_{i};\theta) - \mathbf{Y}_{i} \rvert\rvert ^2_2.
\end{gather}
The naive denoising model learns how to remove noise generated from our pre-defined noise distribution, e.g., Gaussian distribution. This is effective if the real noise follows the same noise distribution. However, in practice, the noise distribution is often much more complex and cannot be simply approximated by predefined patterns. As we will theoretically show in Section~\ref{sec:theory}, the naive denoising method becomes suboptimal when the real noise distribution deviates from our pre-defined noise distributions. Therefore, we incorporate physics constraint to enhance denoising performance, enabling the model to adapt to more complex, real-world noise distributions. \smallskip

\noindent \textbf{Physics-Informed Denoising Model.} Apart from the reconstruction loss in the naive denoising model, we incorporate another \textit{physics-based loss} to overcome the limitation of the naive model. Recall in Section~\ref{sec:def} (Problem Formulation) we mentioned that ground truth clean data satisfy physics constraint $g(\mathbf{x}_{i1},\mathbf{x}_{i2},\cdots,\mathbf{x}_{ic})=0, i=1,2,\cdots,N$. In an ideal scenario, if our denoising model can perfectly eliminate all noise, then the resultant output of this optimal denoising model $f(\cdot;\theta^*)$ will also adhere to the physics constraint:
\begin{equation}
g(f(\mathbf{z}_{i1};\theta^*),f(\mathbf{z}_{i2};\theta^*),\cdots,f(\mathbf{z}_{ic};\theta^*))=0, i=1,2,\cdots,N,
\end{equation}
which can be simplified as $g(f(\mathbf{Z}_{i};\theta^*)) = 0$ or $g(f(\mathbf{Y}_{i}+\mathbf{n}_{i};\theta^*)) = 0$. Therefore, we can exploit the physics constraints intrinsic to an optimal denoising model and formulate the constrained optimization problem as 
\begin{equation}
    \begin{split}
        & \mathrm{argmin}_{\theta} \mathbb{E}_{\mathbf{Y}_{i} \sim \mathcal{Y}, \mathbf{n}_{i} \sim \mathcal{N}}\ell_{\mathrm{rec}}(f(\mathbf{Y}_{i} + \mathbf{n}_{i};\theta),\mathbf{Y}_{i}), \\
    & \mathrm{s.t.} \  g(f(\mathbf{Y}_{i}+\mathbf{n}_{i};\theta)) = 0.
    \end{split}
\end{equation}
Enforcing the physics equation as strict constraints makes the model difficult to optimize. Instead, we include the constraint as a \textit{soft regularization term in the loss function}. Specifically, the loss function for the physics-informed denoising model is a combination of reconstruction loss and physics-based loss:
\begin{gather}
    \mathrm{argmin}_{\theta} \mathbb{E}_{\mathbf{Y}_{i} \sim \mathcal{Y}, \mathbf{n}_{i} \sim \mathcal{N}}\ell(f(\mathbf{Y}_{i} + \mathbf{n}_{i};\theta),\mathbf{Y}_{i}), \\
    \ell = \ell_{\mathrm{rec}} + \lambda \ell_{\mathrm{phy}}, \\
    \ell_{\mathrm{phy}}(f(\mathbf{Y}_{i} + \mathbf{n}_{i};\theta)) =  \lvert\lvert g(f(\mathbf{Y}_{i} + \mathbf{n}_{i};\theta)) \rvert\rvert ^2_2.
\end{gather}
The physics-based loss offers additional prior knowledge about the ground truth data and better guides the learning process of the denoising model, especially when we don't have access to the ground truth distributions. We find in the experiment section that the physics constraint works for both first-order terms and higher-order terms. Here, $\lambda$ is the ratio that balances the two losses. We find that an adaptive balancing strategy that keeps these two losses always at the same orders of magnitude during training is empirically more effective than setting a fixed ratio, as we will illustrate in Section~\ref{sec:sensitivity}. \smallskip

\noindent \textbf{\our Framework.} We design a two-phase training framework for the final denoising model. Specifically, we first optimize the model only with reconstruction loss as a pre-training phase. The two-stage training ensures that training is not biased by the physics loss at the beginning, as the initial exclusive focus on optimizing the physics loss can lead to overly simplistic and trivial solutions, such as all-zero predictions. Two-stage training mitigates this by serving as a warm-up period and providing better denoising model initialization. By learning to reconstruct the data, the model better adapts to the underlying data distribution. In the second phase, we combine both reconstruction loss and physics-based loss to optimize the model. The physics-based loss corrects and guides the model for enhanced denoising performance. To choose the denoising model backbone, we keep both efficacy and efficiency in mind, ensuring that denoising models can run on edge devices with limited compute. Therefore, we use one-dimensional Convolutional Neural Networks (CNN) as an example, as they are relatively lightweight with superior capability in extracting sensor features.

\subsection{Theoretical Analysis}\label{sec:theory}

In this section, inspired by previous works in~\cite{zhussip2019extending,huang2021neighbor2neighbor}, we analytically show that when the inherent noise of our model input has a non-zero mean, training with naive denoising model would be insufficient to recover the ground truth data distribution. 

\begin{corollary}
    Let $\mathbf{y}$ be observed noisy data corrupted from clean data $\mathbf{x}$ by noise $\epsilon$ with non-zero mean $\eta$: $\mathbf{y}=\mathbf{x}+\epsilon, \epsilon \sim \mathcal{E}$. Let $\mathbf{z}$ be manually corrupted data from $\mathbf{y}$ by pre-defined noise $\mathbf{n}$ with zero mean: $\mathbf{z} = \mathbf{y} + \mathbf{n}, \mathbf{n} \sim \mathcal{N}$. Assume $\mathbb{E}_{\mathbf{z}|\mathbf{x}}(\mathbf{z}) = \mathbb{E}_{\mathbf{y}|\mathbf{x}}(\mathbf{y}) = \mathbf{x} + \mathbf{\eta}$, and the variance of $\mathbf{y}$ is $\sigma^2_{\mathbf{y}}$. Then the following equation holds true:
    \begin{equation} 
    \begin{split}
    & \mathbf{E}_{\mathbf{x} \sim \mathcal{X}, \mathbf{\epsilon} \sim \mathcal{E}, \mathbf{n} \sim \mathcal{N}}\lvert\lvert f(\mathbf{x}+\epsilon+\mathbf{n};\theta) - \mathbf{x} \rvert\rvert ^2_2 \\
     & = \mathbf{E}_{\mathbf{x} \sim \mathcal{X}, \mathbf{\epsilon} \sim \mathcal{E}, \mathbf{n} \sim \mathcal{N}}\lvert\lvert f(\mathbf{x}+\epsilon+\mathbf{n};\theta) - (\mathbf{x}+\epsilon) \rvert\rvert ^2_2  \\
     & - \sigma^2_{\mathbf{y}} + 2\eta \mathbf{E}_{\mathbf{x} \sim \mathcal{X}, \mathbf{\epsilon} \sim \mathcal{E}, \mathbf{n} \sim \mathcal{N}}(f(\mathbf{x} + \epsilon+\mathbf{n};\theta) - \mathbf{x}).
    \end{split}
    \end{equation}
\end{corollary}
Since $\eta \neq 0$, optimizing $\mathbf{E}_{\mathbf{x} \sim \mathcal{X}, \mathbf{\epsilon} \sim \mathcal{E}, \mathbf{n} \sim \mathcal{N}}\lvert\lvert f(\mathbf{x}+\epsilon+\mathbf{n};\theta) - (\mathbf{x}+\epsilon) \rvert\rvert ^2_2$ is not equivalent to the ideal goal of optimizing against ground truth, i.e., $\mathbf{E}_{\mathbf{x} \sim \mathcal{X}, \mathbf{\epsilon} \sim \mathcal{E}, \mathbf{n} \sim \mathcal{N}}\lvert\lvert f(\mathbf{x}+\epsilon+\mathbf{n};\theta) - \mathbf{x} \rvert\rvert ^2_2 $. In summary, the corollary states that the naive denoising model's optimization target fails to completely mitigate the inherent noises. Therefore, we incorporate physics-based loss to augment the denoising performance, particularly when underlying noise distributions are inaccessible.
\section{Evaluation}\label{sec:eval}
\subsection{Experimental Setup}

We use a one-dimensional convolutional neural network with four layers as the backbone for our denoising model. The respective kernel sizes for each layer are set to $7,5,3,3$.
We use Adam optimizer with learning rate $1e^{-4}$ and batch size $16$. For each dataset $\mathcal{D}_X=\{\mathbf{X}_i\}_{i=0}^N$, we first compute how well each sample aligns with the physics equation. Specifically, the alignment for the $i$-th sample is the L2 norm $a_i=\lvert\lvert g(\mathbf{X}_i) \rvert\rvert_2^2$. We select samples with the top 50\% smallest $a_i$ as our training set. These samples align more coherently with physics equations and potentially carry less noise. The remaining samples are used as the test set. 
The applications we have studied are mostly stationary systems, so the model should be able to generalize to new test data in practice.
We conduct the experiments in \textsc{Pytorch} with NVIDIA RTX A6000 (with 48GB memory), AMD EPYC 7452 32-Core Processor, and Ubuntu 18.04.5 LTS. We also implement our method on edge device (Raspberry Pi 4) and evaluate its efficiency in Section~\ref{sec:efficiency}. We tune the hyper-parameters of both \our and baselines to minimize loss on the training set, and then evaluate on the test set after hyper-parameter tuning. 

We compare \our with both statistical methods and recent deep learning denoising methods as follows (including two self-supervised methods DIP and N2N):

\begin{figure*}[t]
    \centering
    \includegraphics[width=\linewidth]{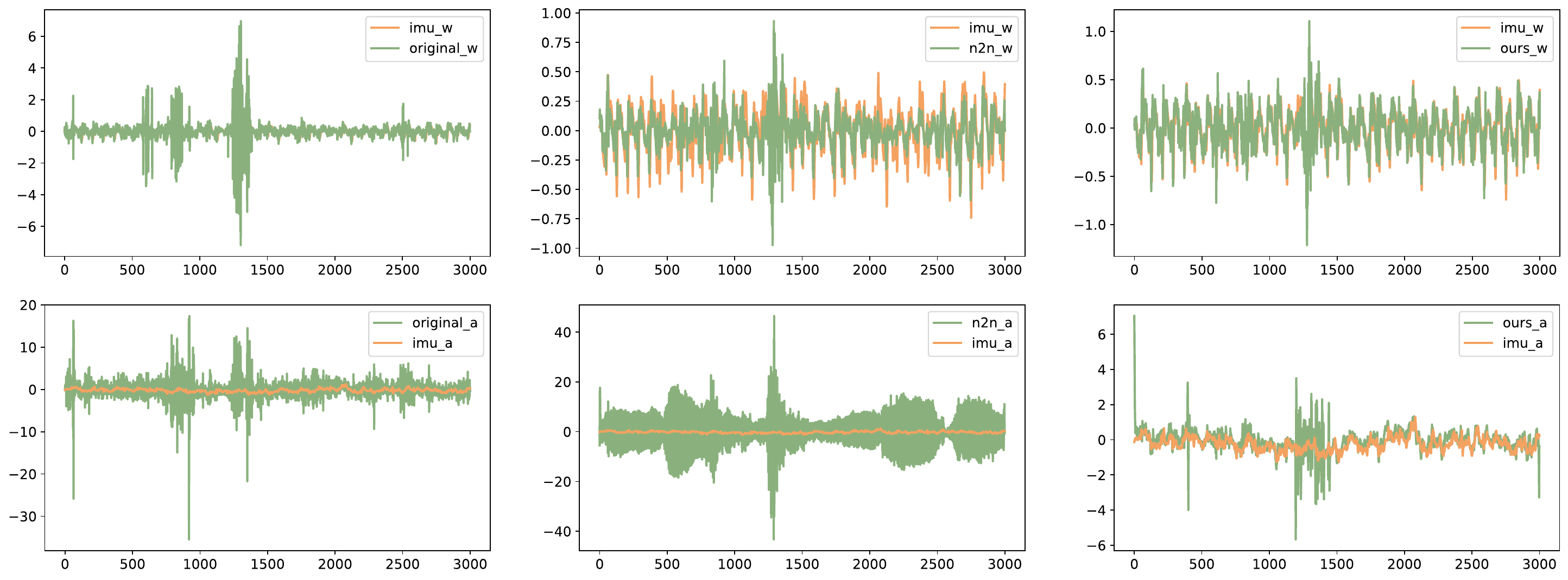}
    \caption{Top: Angular velocity from IMU (imu\_w) and from first-order derivative calculation of orientation based on original data (original\_w)/N2N(n2n\_w)/\our (ours\_w). Bottom: Acceleration from IMU (imu\_a) and from second-order derivative calculation of location based on original data (original\_a)/N2N(n2n\_a)/\our (ours\_a). X-axis: timestep, Y-axis: meter.}
    \label{fig:imu}

      \begin{subfigure}[b]{0.45\textwidth}
          \centering
          \includegraphics[width=\textwidth]{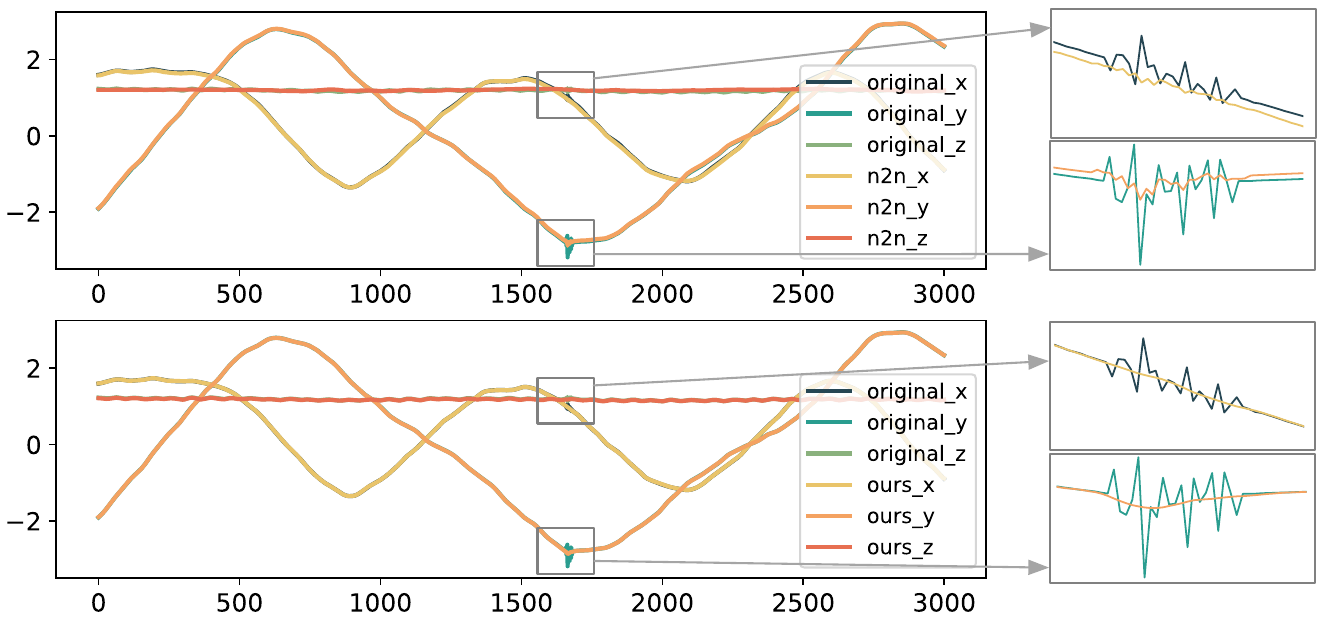}
          \caption{Location (Top: N2N, Bottom: \our)}
          \label{fig:recons_pos}
      \end{subfigure}
      \hspace{0.03\textwidth}
      \begin{subfigure}[b]{0.45\textwidth}
          \centering
          \includegraphics[width=\textwidth]{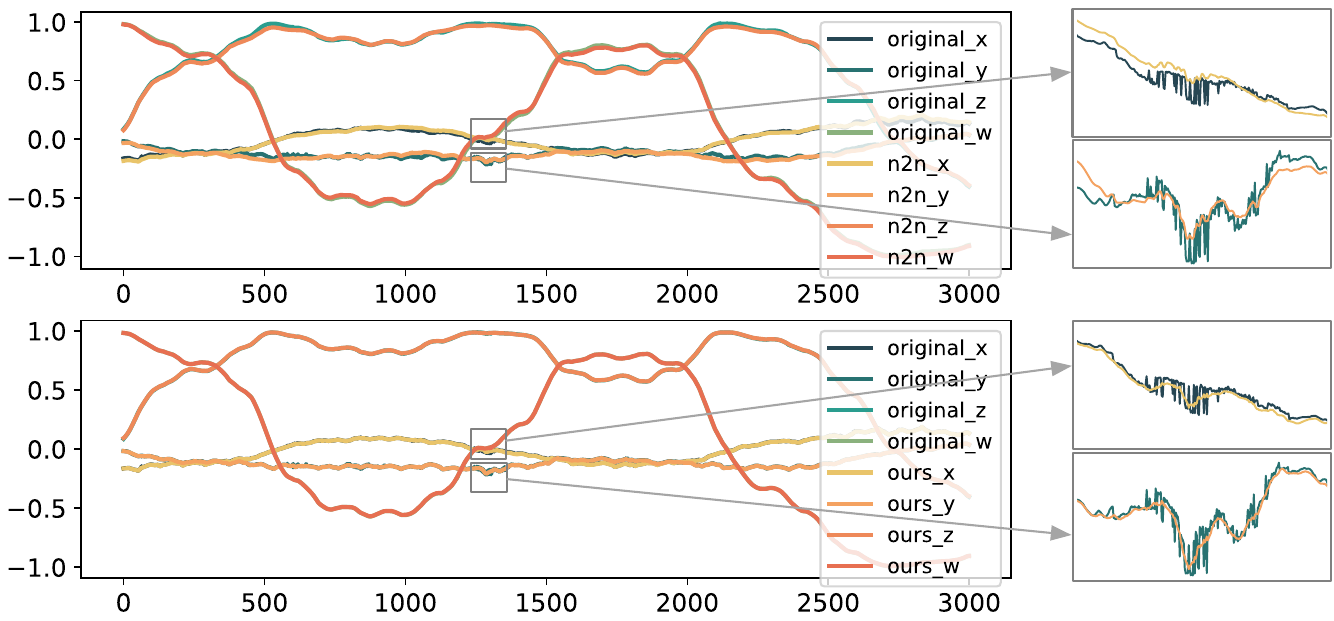}
          \caption{Orientation (Top: N2N, Bottom: \our)}
          \label{fig:recons_quat}
      \end{subfigure}
         \caption{We compare \our with the best performing baseline N2N for qualitative denoising performance of (a) location and (b) orientation. We also zoom in for regions with higher noise levels for better illustration. \our generates smoother results which also align more closely with the original locations and orientations. X-axis: timestep, Y-axis: meter.} 
         \label{fig:recons}
\end{figure*}

\begin{table}[t]
\centering
    \caption{Physics alignment results for OxIOD dataset. We bold the best and \underline{underline} the second best. \our aligns the best with both acceleration and angular velocity.}
	\setlength{\tabcolsep}{2.5mm}{
 \scalebox{0.9}{
    \begin{tabular}{rcccc}
    \toprule
        Model & \multicolumn{2}{c}{Acceleration (m/s$^2$) } & \multicolumn{2}{c}{Angular Velocity (rad/s)} \\
    \cmidrule(lr){2-3} \cmidrule(lr){4-5}

        Metrics & MSE & MAE & MSE & MAE \\
    \midrule
        Original & 762.6 & 3.7862 & 2.6219 & 0.2376 \\
        Gaussian & 363.2 & \underline{3.2295} & 1.6277 & 0.2161  \\
        DWT & 854.9 & 5.4534 & 2.6034 & 0.2701 \\
        DnCNN & 312.5 & 8.3830 & \underline{0.3470} & 0.1896  \\
        TSTNN & 3272.0 & 30.513 & 0.4184 & 0.4836 \\
        DIP & 2153.6 & 33.938 & 0.3788 & 0.4013   \\
        N2N & \underline{118.7} & 4.5749 & 0.3565 & \underline{0.1756} \\
    \midrule
        \our & \textbf{1.8695} & \textbf{0.6372} &\textbf{0.0380} & \textbf{0.0690}  \\
    \bottomrule
    \end{tabular}
    }
    }
    \label{tab:oxiod-recons}
\end{table}

\begin{itemize}[nosep,leftmargin=*]
    \item  \noindent \textbf{Gaussian Filter} employs a Gaussian filter with standard deviation $\sigma$ for the kernel to smooth the sensor data.
    \item \noindent \textbf{DWT}~\cite{de2017insights} decomposes the signal using Discrete Wavelet Transform (DWT) and chooses the most energetic coefficients to reconstruct the denoised signal.
    \item \noindent \textbf{DnCNN}~\cite{zhang2017beyond} is a convolutional neural network-based denoising model that exploits residual learning and batch normalization to boost the denoising performance. 
    \item \noindent \textbf{TSTNN}~\cite{wang2021tstnn} is a Transformer-based model for end-to-end speech denoising. It is composed of a feature-mapping encoder, a two-stage Transformer model to extract local and global information, a masking module and a decoder to reconstruct denoised speech.
    \item \noindent \textbf{Deep Image Prior (DIP)}~\cite{ulyanov2018deep} exploits the structure of the neural network as priors for denoising data. The method fits the model to the noisy data with early exit, so the network captures dominant patterns while ignoring the random noise. 
    \item \noindent \textbf{Neighbor2Neighbor (N2N)}~\cite{huang2021neighbor2neighbor} applies random neighbor sub-sampler on noisy data to generate input and target for the network. It also proposes a regularizer as additional loss for performance augmentation.
\end{itemize}

We evaluate the performance from three perspectives. First of all, if we have collected data of higher quality using other sources (e.g., more expensive and accurate sensors), we would regard these higher-quality data as approximate ground truth and compute Mean Square Error (MSE) and Mean Absolute Error (MAE) between the denoised data $\mathbf{\hat{X}}$ and the higher-quality data $\mathbf{{X}}$ as the first metric, noted as ``Reconstruction Performance''. Specifically, Reconstruction MSE$\ =\lvert\lvert \mathbf{\hat{X}}-\mathbf{{X}} \rvert\rvert^2_2$, and Reconstruction MAE $\ = \lvert\lvert \mathbf{\hat{X}}-\mathbf{{X}} \rvert\rvert_1$. The second metric investigates alignment of denoised data with the governing physics equations, noted as ``Physics Alignment''. Specifically, $\mathrm{Physics \ MSE} = \lvert\lvert g(\mathbf{\hat{X}}) \rvert\rvert^2_2$, and Physics MAE$\ = \lvert\lvert g(\mathbf{\hat{X}}) \rvert\rvert_1$. Lastly, if the denoised data are further applied in downstream applications (e.g., inertial navigation system), we also evaluate ``Downstream Performance'' based on the metrics of interest for the corresponding downstream task. Measuring downstream performance helps us understand the full impact of denoising on real-world applications.

\subsection{Application I: Inertial Navigation}

\subsubsection{Dataset}
We use OxIOD~\cite{chen2018oxiod} dataset for inertial navigation experiments. OxIOD collects accelerometer and gyroscope data (100 Hz) mostly by IMUs (InvenSense ICM20600) in iPhone 7 plus. A Vicon motion capture system (10 Bonita B10 cameras) is used to record the locations and orientations. The total walking distance and recording time of the dataset are 42.5 km and 14.72 h. The data collection protocol is designed to simulate natural pedestrian movement within an indoor environment equipped with motion capture system. When collecting the data, a pedestrian walks naturally inside a room with motion capture system, carrying the phone in hand, in the pocket, in the handbag, or on the trolley. 

\begin{table*}[t]
\centering
    \caption{Inertial navigation performance on OxIOD dataset with two inertial navigation models (IONet and RoNIN). We bold the best and \underline{underline} the second best. When applying the denoised data to downstream inertial navigation task, \our yields the best performance given two different inertial navigation models (IONet and RoNIN).}
 \scalebox{0.9}{
    \begin{tabular}{rcccccccccccc}
    \toprule
        Model & \multicolumn{6}{c}{IONet~\cite{chen2018ionet}} & \multicolumn{6}{c}{RoNIN~\cite{herath2020ronin}} \\
    \cmidrule(lr){2-7} \cmidrule(lr){8-13}

        \multirow{2}{*}{Metrics} & vx & vy & vz & mean v & ATE & RTE & vx  & vy & vz & mean v  & ATE & RTE  \\
        & (m/s) & (m/s) & (m/s) & (m/s) & (m) & (m) & (m/s) & (m/s) & (m/s) & (m/s) & (m) & (m) \\
    \midrule
        Original & 0.0207 & 0.0642 & \underline{0.0093} & 0.0314 & 0.3076 & 0.8194  & 0.0180 & 0.0621 & \underline{0.0090} & 0.0297 & \underline{0.2472} & \underline{0.6337} \\
        Gaussian & 0.0249 & 0.0496 & 0.0145 & 0.0297 & 0.6111 & 1.8727  & 0.0242 & 0.0498 & 0.0147 & 0.0296 & 0.5988 & 1.8427   \\
        DWT & 0.0266 & 0.0732 & 0.0094 & 0.0364 & 0.3142 & 0.8079 & 0.0243 & 0.0714 & 0.0091 & 0.0349 & 0.2665 & 0.7023 \\ 
        DnCNN & \underline{0.0200} & 0.0235 & 0.0144 & 0.0193 & \underline{0.3001} & \underline{0.7891} & \underline{0.0177} & 0.0213 & 0.0139 & \underline{0.0176} & 0.2476 & 0.6598   \\
        TSTNN & 0.2857 & 0.3250 & 0.0935 & 0.2348 & 0.6496 & 1.6575 & 0.2865 & 0.3253 & 0.0938 & 0.2352 & 0.6256 & 1.5794 \\
        DIP & 0.1971 & 0.2650 & 0.0105 & 0.1576 & 0.5759 & 1.5108  & 0.1926 & 0.2570 & 0.0101 & 0.1533 & 0.3989 & 1.0358   \\
        N2N & 0.0246 & \underline{0.0144} & 0.0183 & \underline{0.0191} & 0.3151 & 0.8317 & 0.0224 & \underline{0.0122} & 0.0182 & \underline{0.0176} & 0.2605 & 0.6956 \\
    \midrule
        \our & \textbf{0.0102} & \textbf{0.0095} & \textbf{0.0031} & \textbf{0.0076} & \textbf{0.2998} & \textbf{0.7875} & \textbf{0.0081} & \textbf{0.0078} & \textbf{0.0017} & \textbf{0.0059} & \textbf{0.2413} & \textbf{0.6309}  \\
    \bottomrule
    \end{tabular}
    }
    \label{tab:ionet}
\end{table*}

\begin{figure*}[t]
      \centering
      \begin{subfigure}[b]{0.45\textwidth}
          \centering
          \includegraphics[width=\textwidth]{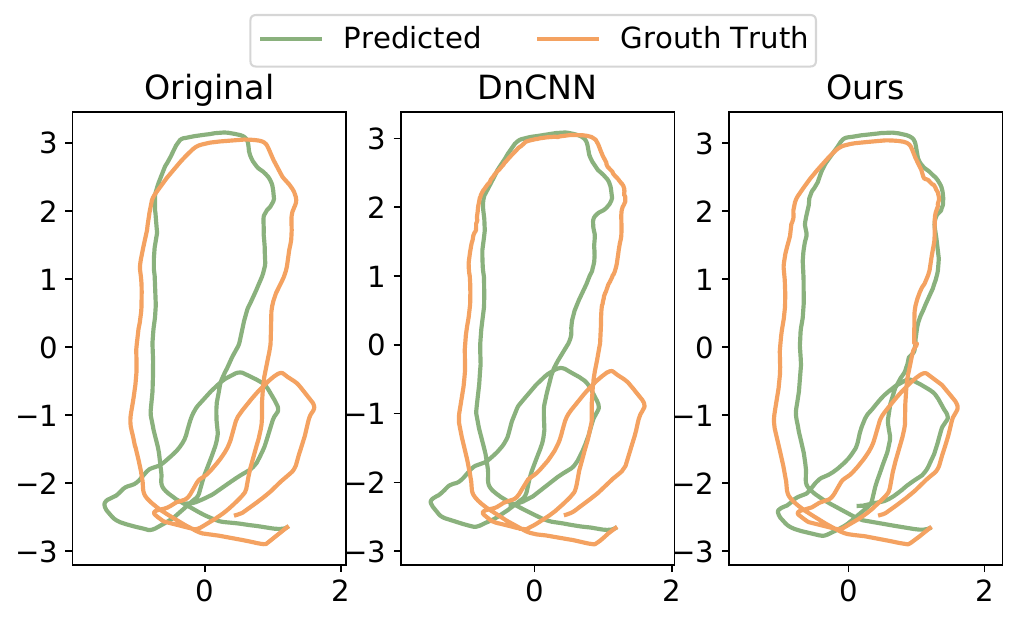}
          \caption{Case 1}
          \label{fig:ronin_case1}
      \end{subfigure}
      \hspace{0.03\textwidth}
      \begin{subfigure}[b]{0.45\textwidth}
          \centering
          \includegraphics[width=\textwidth]{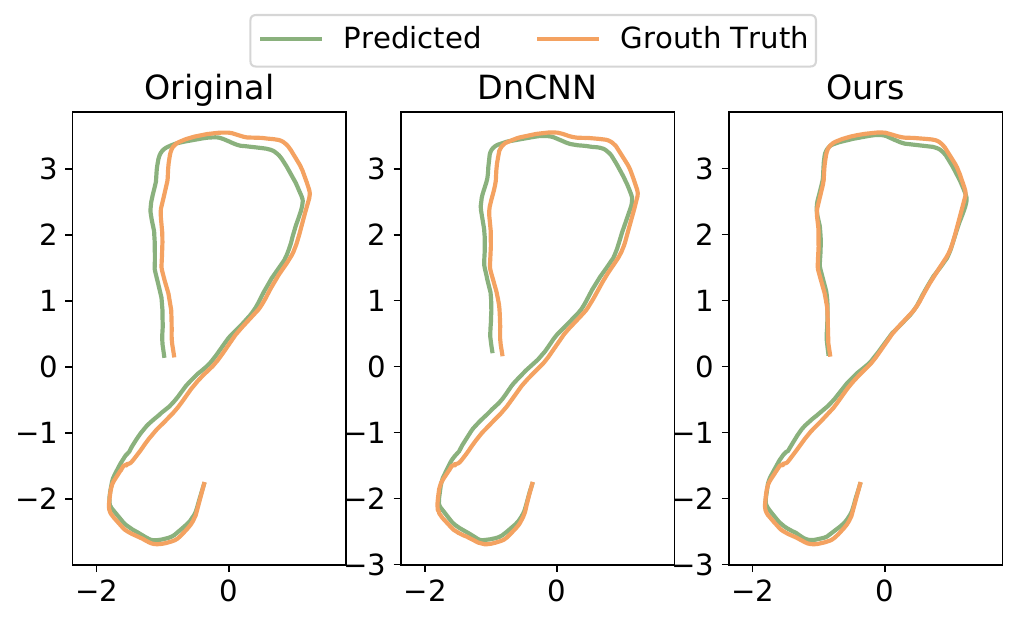}
          \caption{Case 2}
          \label{fig:ronin_case2}
      \end{subfigure}
         \caption{Downstream inertial navigation performance on OxIOD dataset with RoNIN model. RoNIN trained on denoised data of \our reconstructs trajectories closet to the ground truth trajectories. X-axis: meter, Y-axis: meter.} 
         \label{fig:ronin}
\end{figure*}

\subsubsection{Physics Alignment}
We compare the physics alignment performance in Table~\ref{tab:oxiod-recons}. Specifically for this application, we compute the misalignment for acceleration and angular velocity. Adapting our experimental approach for the specificities of inertial navigation, we only denoise the location and orientation data, and leave IMU data in its original form. This is because the downstream inertial navigation model's training requires integration over the IMU, thereby inherently smoothing and denoising the IMU data. Since our model only denoises the location and orientation data, we model the physics relationships between the denoised location, orientation, and the original IMU. More specifically, acceleration MSE in Table~\ref{tab:oxiod-recons} is computed as the discrepancy between accelerometer data from IMU ($a$) and the second-order derivative of the denoised location subtracted by gravity constant and multiplied by rotation matrix ($R_q^T ({d^2p}/{dt^2} - g_0)$), as shown in Equation~\ref{eq:accel}. Similarly, angular velocity MSE is computed as the difference between gyroscope data from IMU multiplied by orientation ($q \otimes w / 2$) and the first-order derivative of denoised orientation (${dq}/{dt}$), as in Equation~\ref{eq:quat}. 

From Table~\ref{tab:oxiod-recons}, we can see that while most denoising methods reduce the physics MSE compared with original data without denoising, \emph{\our significantly reduces the physics MSE by two orders of magnitude}. Accompanied by these quantitative results, we also offer qualitative illustrations in Figure~\ref{fig:imu}. We present a visual comparison of the direct IMU measurements and derived accelerations and angular velocities. When comparing \our with original and denoised data from the best-performing baseline, we observe that \our-denoised data provide the best physics alignment for both angular velocity and acceleration. 

\subsubsection{Reconstruction Performance}

For the inertial navigation application, we don't have access to ground truth location and orientation data. Therefore, we qualitatively compare the reconstruction performance of \our and the best-performing baseline in Figure~\ref{fig:recons}. We compare the reconstruction performance for both location and orientation, with detailed zoom-ins on regions exhibiting higher noise levels for more clear illustration. We observe that \our effectively reduces the noise in the original data, while simultaneously reconstructing results that show greater alignment and coherence with the original locations and orientations.

\begin{figure*}[t]
      \centering
      \begin{subfigure}[b]{0.23\textwidth}
          \centering
          \includegraphics[width=\textwidth]{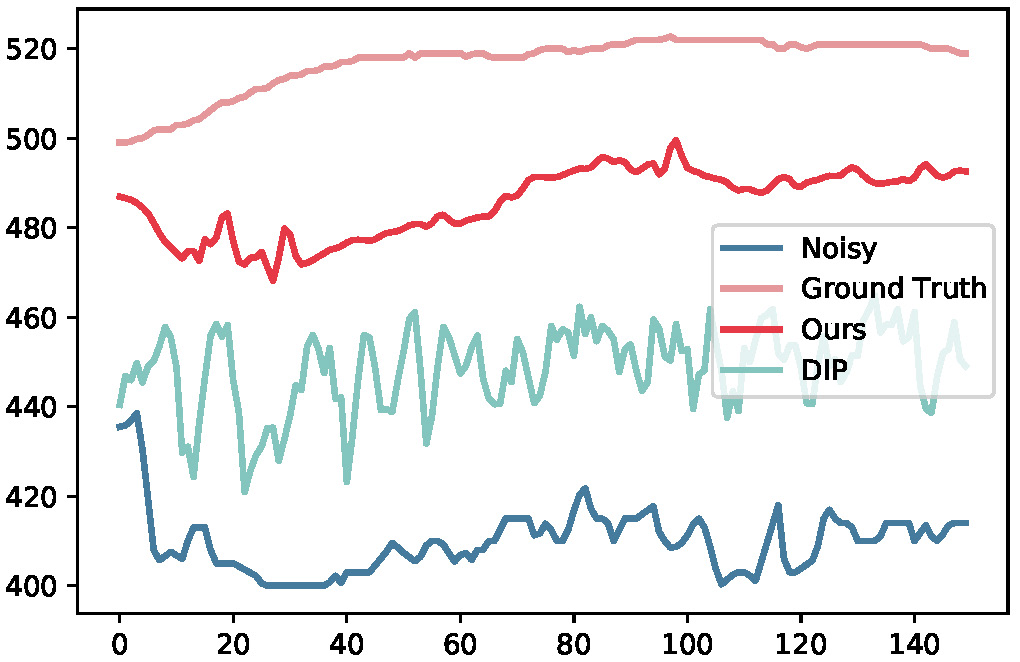}
          \caption{Example 1}
          \label{fig:co2_case1}
      \end{subfigure}
      \begin{subfigure}[b]{0.23\textwidth}
          \centering
          \includegraphics[width=\textwidth]{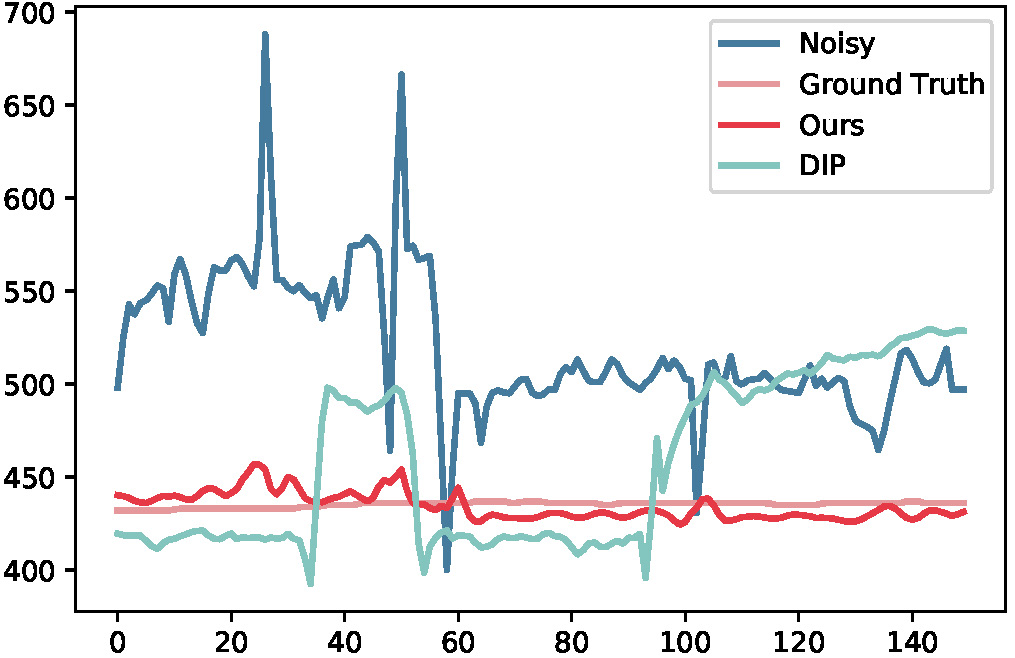}
          \caption{Example 2}
          \label{fig:co2_case2}
      \end{subfigure}
      \begin{subfigure}[b]{0.23\textwidth}
          \centering
          \includegraphics[width=\textwidth]{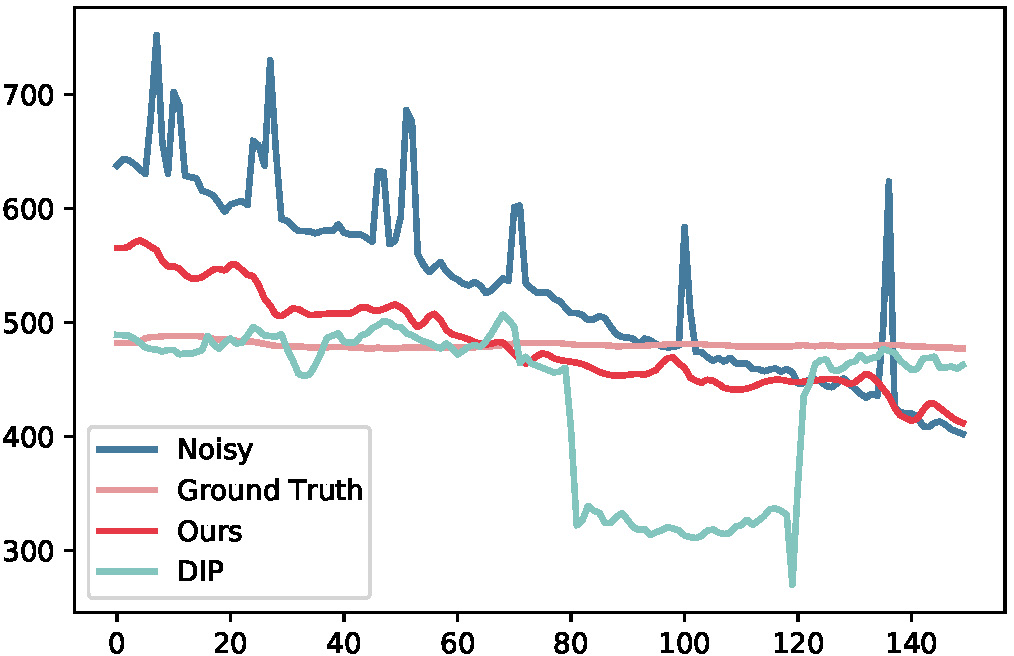}
          \caption{Example 3}
          \label{fig:co2_case3}
      \end{subfigure}
      \begin{subfigure}[b]{0.23\textwidth}
          \centering
          \includegraphics[width=\textwidth]{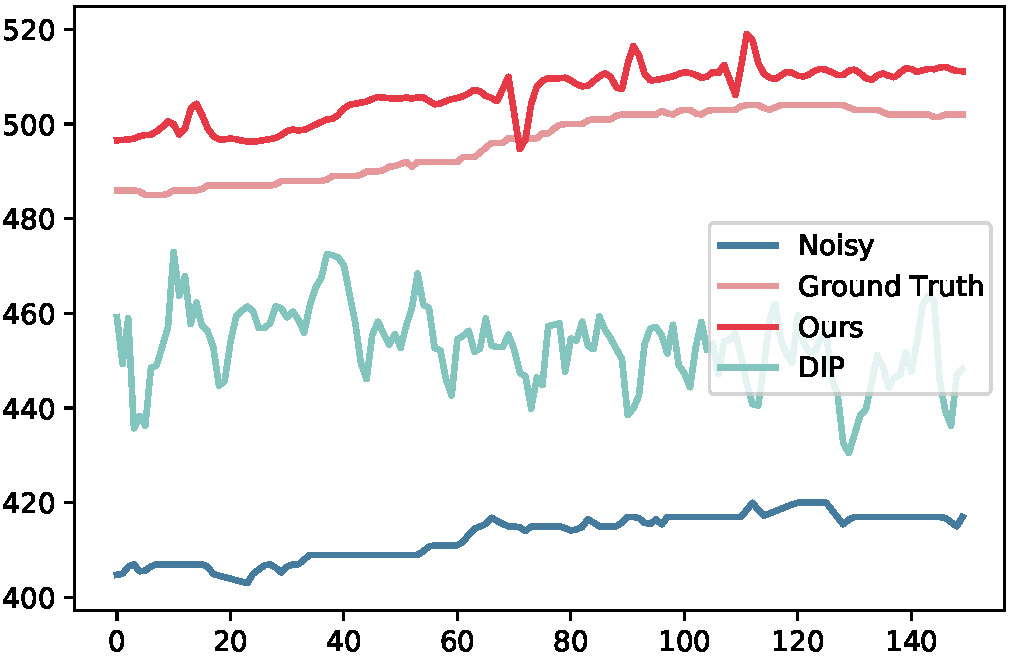}
          \caption{Example 4}
          \label{fig:co2_case4}
      \end{subfigure}
         \caption{We compare the CO$_2$ denoising results of \our and the best performing baseline DIP. \our denoises CCS881 results closet to the ground truth (K30 measurements). X-axis: timestep, Y-axis: ppm.} 
         \label{fig:CO2_case}
\end{figure*}

\begin{figure}[b]
      \centering
      \begin{subfigure}[b]{0.23\textwidth}
          \centering
          \includegraphics[width=\textwidth]{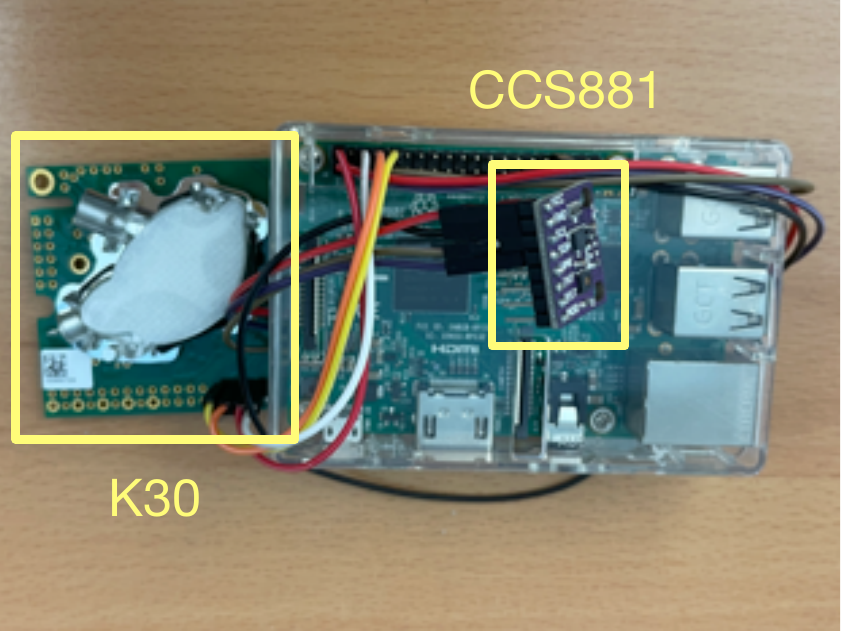}
          \caption{Deployment on desk}
          \label{fig:sensor}
      \end{subfigure}
      \begin{subfigure}[b]{0.23\textwidth}
          \centering
          \includegraphics[width=\textwidth]{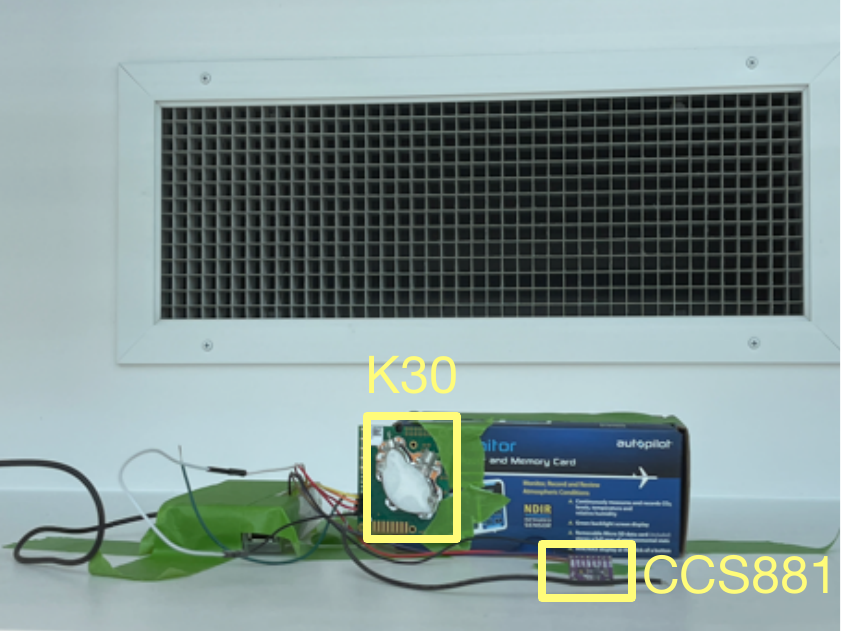}
          \caption{Deployment by vent}
          \label{fig:deployment}
      \end{subfigure}
         \caption{We connected a pair of two different CO$_2$ sensors (K30 and CCS 881 respectively) on a Rasberry Pi. We deployed (a) one pair on the desk in the center of the office  and (b) the other pair close to the exhaust vent to monitor CO$_2$ level.} 
         \label{fig:device}
\end{figure}

\subsubsection{Downstream Performance}
We also use the denoised data to train downstream inertial navigation model, and compare \our and baselines in terms of downstream task performance. We use IMU data, denoised location and orientation data as training pairs for the following inertial navigation models:

\begin{itemize}[nosep,leftmargin=*]

\item \noindent \textbf{IONet}~\cite{chen2018ionet}: IONet uses deep recurrent neural networks to learn location transforms in polar coordinates from raw IMU data, and construct inertial odometry.

\item \noindent \textbf{RoNIN}~\cite{herath2020ronin}: RoNIN regresses velocity given IMU with coordinate frame normalization and robust velocity losses.
\end{itemize}

We measure the downstream performance by the following three metrics, following previous works~\cite{herath2020ronin}.

\begin{itemize}[nosep,leftmargin=*]
\item \noindent \textbf{Velocity}: We compute MSE of predicted velocity (x, y, and z-axis, along with the mean velocity of all three axes).

\item \noindent \textbf{Absolute Translation Error (ATE)}: ATE is defined as the Root Mean Squared Error (RMSE) between the entire estimated and ground truth location trajectory. 

\item \noindent \textbf{Relative Translation Error (RTE)}: RTE is defined as the Root Mean Squared Error (RMSE) over a fixed time interval. We choose 1 minute in our experiment as in previous works~\cite{herath2020ronin}.
\end{itemize}

We compare \our with baselines on downstream inertial navigation task using IONet and RoNIN (Table~\ref{tab:ionet}) as the downstream models. \emph{We can observe that \our best predicts the velocity and locations, regardless of the specific inertial navigation model.} In Figure~\ref{fig:ronin}, we also qualitatively compare \our with the best-performing baseline using RoNIN as the inertial navigation model. We can see \our's denoised data offer RoNIN the best trajectory reconstruction performance relative to both the original data and that from the best-performing baseline.

\subsection{Application II: CO$_2$ Monitoring}\label{sec:co2}

\subsubsection{Dataset Collection} In this experiment, we deployed two pairs of CO$_2$ sensors (K30 and CCS811) in a typical graduate student office environment. K30 is a highly accurate Non-Dispersive Infra-Red (NDIR) CO$_2$ sensor manufactured by SenseAir which costs \$100 per unit. It features an accuracy of $\pm 30$ ppm or $\pm 3\%$ of measurements and a high repeatability of $\pm 20$ ppm or $\pm 1\%$ of measurements~\cite{K30}. The NDIR technology~\cite{Palzer2020-su} utilizes the unique property of CO$_2$ molecules --- their significant absorption of infrared (IR) light in the vicinity of 4.2 $\mu$m wavelength. When a gas sample is illuminated with light of this particular wavelength, the concentration of CO$_2$ can be deduced by examining the fraction of light absorbed. We regard K30 readings as the ground truth of CO$_2$ concentration.

In comparison, CCS811 \cite{CCS811} is a substantially cheaper Metal-Oxide (MOX) gas sensor, priced at less than \$10 per unit, which computes equivalent CO$_2$ readings based on hydrogen gas readings. A MOX gas sensor works by measuring and analyzing changes in the conductivity of the gas-sensitive MOX semiconductor layer(s) at various gas exposure \cite{MOX}. It has no verifiable accuracy or repeatability guarantee and is usually not recommended for laboratory use. 

For each pair of these two sensors, we connect both of them to a single Raspberry Pi board, which collects data through UART protocol and I2C interface from K30 and CCS811 respectively, at an interval of 4 seconds. The collected readings are transmitted to a remote InfluxDB instance for storage and further analysis. We placed these two pairs of CO$_2$ sensors as follows. One pair is adjacent to the HVAC exhaust vent and the other pair is at the center of the room, to capture the relevant variables in Equation~\ref{eq:co2} as illustrated in Figure~\ref{fig:device}. We set inflow CO$_2$ concentration in Equation~\ref{eq:co2} to 440 ppm based on empirical observation. 

\begin{table}[t]
\centering
    \caption{Reconstruction and physics alignment results on CO$_2$ dataset. We bold the best and \underline{underline} the second best.} 
	\scalebox{0.9}{
 {\setlength{\tabcolsep}{3mm}
    \begin{tabular}{rcccc}
    \toprule
    Model & \multicolumn{2}{c}{Recons ($1\times10^6$ ppm)} & \multicolumn{2}{c}{Physics ($1\times10^6$ ppm)} \\
    \cmidrule(lr){2-3} \cmidrule(lr){4-5}
        Metrics & MSE & MAE &  MSE & MAE  \\
    \midrule
        Original & 1.5654 & 0.0020 & 0.1082 & 0.1908 \\
        Gaussian & 0.6265 & 0.0016 & 0.0278 & 0.1019   \\
        DWT & 1.5076 & 0.0020 & 0.1045 & 0.1846 \\
        DnCNN & 1.5381 & 0.0020 & 0.1064 & 0.1897  \\
        TSTNN & 0.0956 & 0.0007 & 0.0027 & 0.0498 \\
        DIP & \underline{0.0841} & \underline{0.0006} & \underline{0.0018} & \underline{0.0308} \\
        N2N & 0.4396 & 0.0018 & 0.0085 & 0.0789 \\
    \midrule
        \our & \textbf{0.0371} & \textbf{0.0004} & \textbf{0.0012} & \textbf{0.0200} \\
    \bottomrule
    \end{tabular}}
    }
    \label{tab:co2-recons}
\end{table}

\subsubsection{Reconstruction and Physics Performance}
For the application of CO$_2$ monitoring, data collected from high-precision K30 sensors serve as the benchmark for calculating the reconstruction performance. We also evaluate the physics alignment by how well our denoised data match Equation~\ref{eq:co2}. As shown in Table~\ref{tab:co2-recons}, \emph{\our outperforms all baselines, showcasing the lowest reconstruction and physics alignment errors}. We also qualitatively compare \our with the best-performing baseline in Figure~\ref{fig:CO2_case}. We can observe that the denoised data by \our align the closest to the ground truth (K30 data) compared with original noisy data as well denoised data from the best-performing baseline.  

\begin{figure}[b]
      \centering
      \begin{subfigure}[b]{0.23\textwidth}
          \centering
          \includegraphics[width=\textwidth]{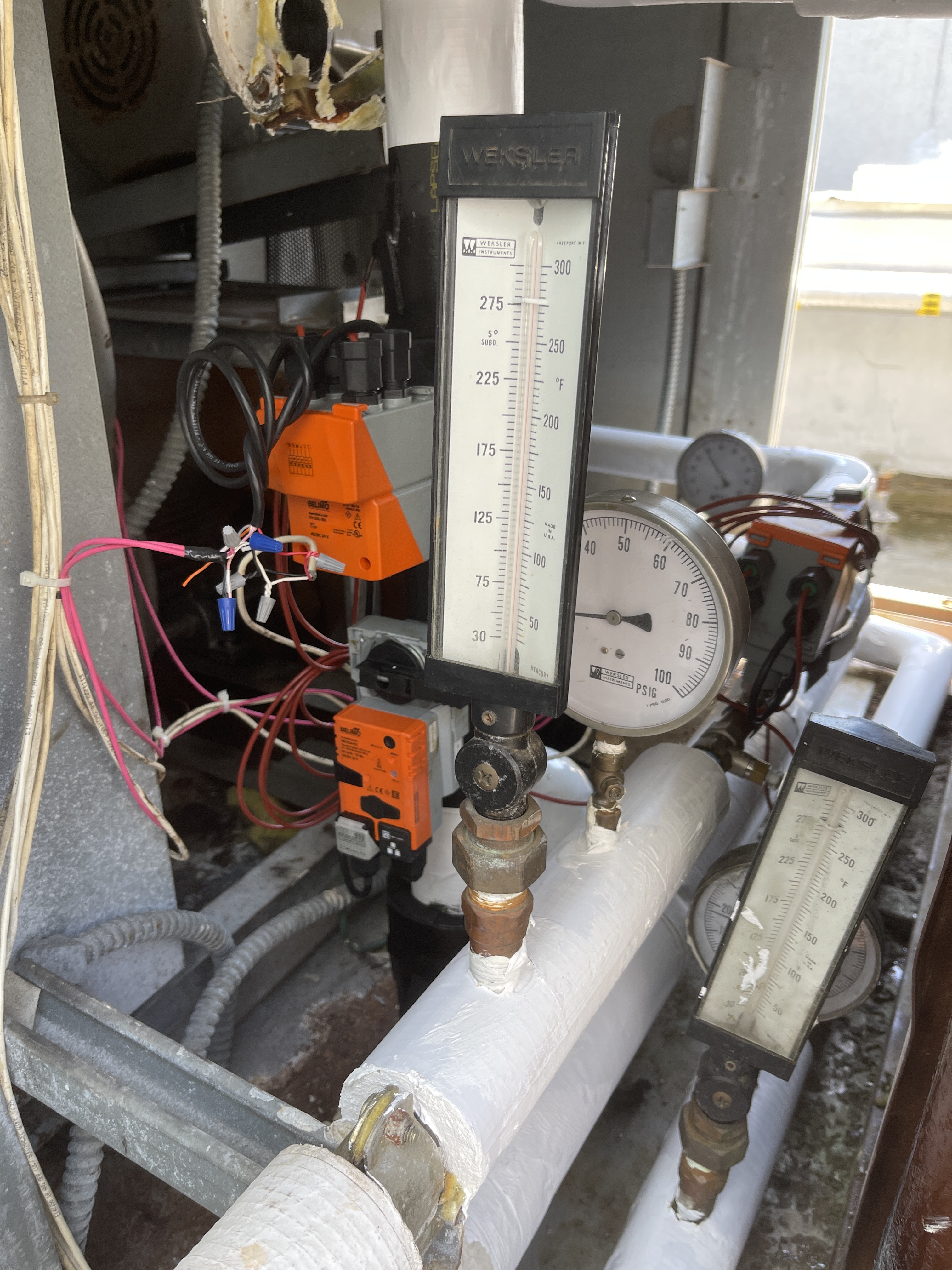}
          \caption{$\Delta Q$}
          \label{fig:hvac-dq}
      \end{subfigure}
      \begin{subfigure}[b]{0.23\textwidth}
          \centering
          \includegraphics[width=\textwidth]{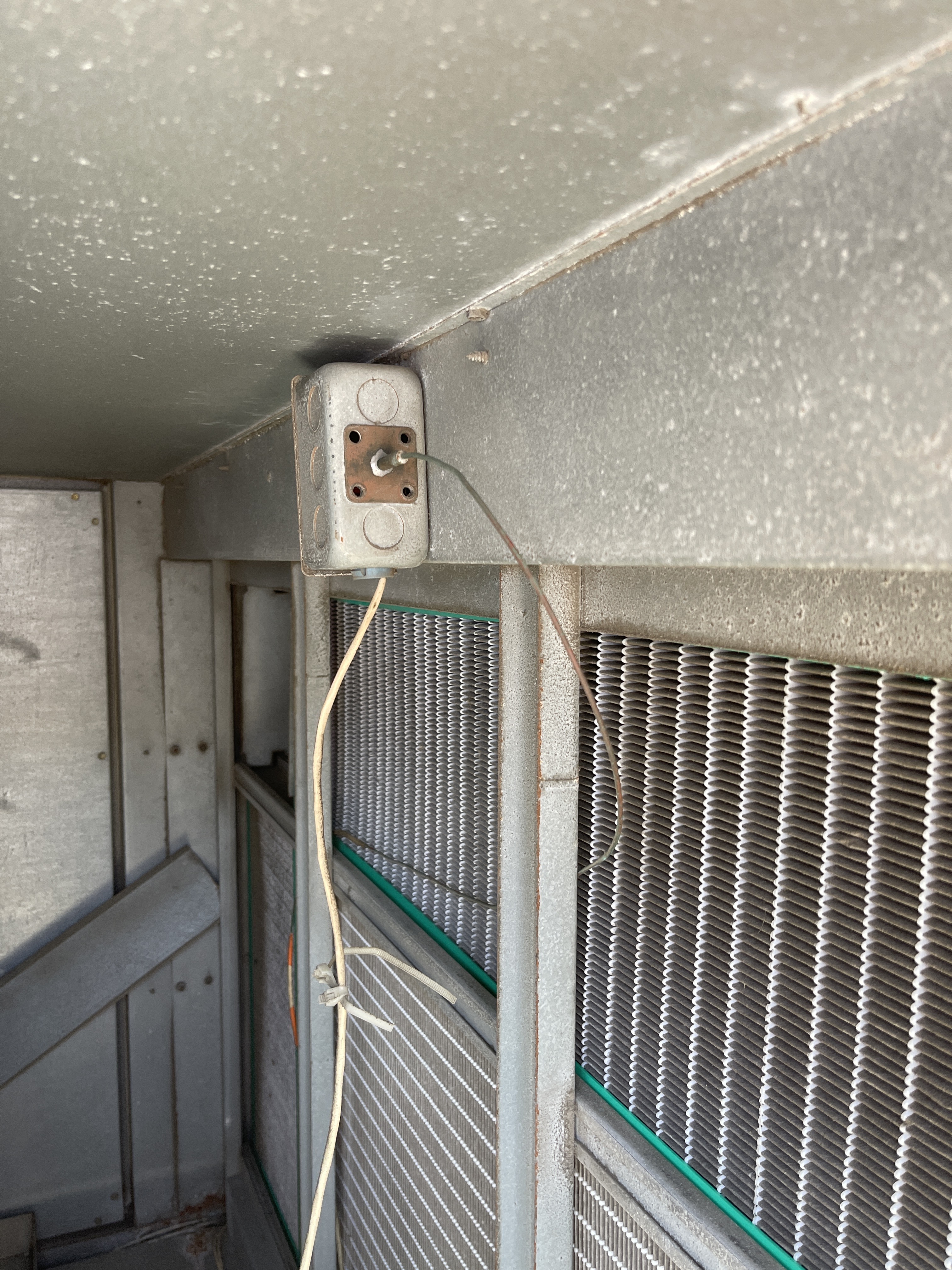}
          \caption{$\Delta T$}
          \label{fig:hvac-dt}
      \end{subfigure}
         \caption{Sensor deployments in HVAC systems. (a) We measure heat difference $\Delta Q$ through the water temperature difference. (b) We measure the air temperature differences with a copper thread averaging sensor. 
         }
         \label{fig:hvac-device}
\end{figure}

\subsection{Application III: Air Handling in HVAC Control System}\label{sec:hvac}

\subsubsection{Dataset Collection} We collected data points of relevant variables in Equation~\ref{eq:hvac} from an AHU serving a lecture hall on campus. We chose this particular AHU based on the following two criteria. Firstly, this AHU is equipped with both heating and cooling thermal submeters (Figure~\ref{fig:hvac-dq}), ensuring the availability of $\Delta Q$ in Equation~\ref{eq:hvac}. Secondly, none of the zones served by this AHU has any reheat unit. In theory, this guarantees that our assumption holds true that heating and cooling coils are the only heat sources. We collected the data through the campus building management system (BMS) for 18 consecutive days, with readings taken at 1-second intervals. We plan to collect more long-term data as a future work.

The mixed air temperature and supply air temperature are measured by copper-thread averaging duct sensors (costing \$150 for each) within the respective supply and exhaust ducts of this AHU, as shown in Figure~\ref{fig:hvac-dt}. These sensors give the average temperature based on the fact that the electrical resistance of copper changes in a predictable way with temperature changes. However, this method is prone to inaccuracy and lacks repeatability due to challenges associated with maintenance and cleaning, as well as the uneven distribution of the thread within the duct. A feasible yet costly alternative solution is to use multiple single-unit temperature sensors evenly distributed throughout the duct. These sensors are more expensive and harder to deploy. To evaluate the denoising methods, we carried this setup by deploying two dual-probe high-accuracy, industry-level temperature sensors in the supply duct, each costing \$200, and took the average of readings from each probe to get a highly reliable ground truth temperature measurement. 

\begin{table}[t]
\centering
    \caption{Reconstruction and physics alignment on HVAC dataset. We bold the best and \underline{underline} the second best. \our best denoises HVAC data in terms of both reconstruction accuracy as well as physics alignment.}
	\scalebox{0.9}{
 \setlength{\tabcolsep}{3mm}{
    \begin{tabular}{rcccc}
    \toprule
    Model & \multicolumn{2}{c}{Reconstruction (K)} & \multicolumn{2}{c}{Physics (K)} \\
    \cmidrule(lr){2-3} \cmidrule(lr){4-5}
        Metrics & MSE & MAE &  MSE & MAE  \\
    \midrule
        Original & 0.9479 & 0.8841 & 50.302 & 6.7184 \\
        Gaussian & 0.8687 & 0.8782 & 49.528 & 6.7003  \\
        DWT &  0.8553 &0.8689 & 48.782 & 6.6471 \\
        DnCNN & \underline{0.3284} & \underline{0.4786} & 50.380 & 6.7241  \\
        TSTNN & 2.2980 & 1.1980 & 32.012 & 3.7260 \\
        DIP & 3.7336 & 1.5098 & \underline{29.747} & 3.7410 \\
        N2N & 1.1830 & 0.9522 & 31.748 & \underline{3.6990} \\
    \midrule
        \our & \textbf{0.1994} & \textbf{0.3454} & \textbf{14.081} & \textbf{3.1600} \\
    \bottomrule
    \end{tabular}}
    }
    \label{tab:hvac}
\end{table}

\begin{figure}[t]
      \centering
      \begin{subfigure}[b]{0.24\textwidth}
          \centering
          \includegraphics[width=\textwidth]{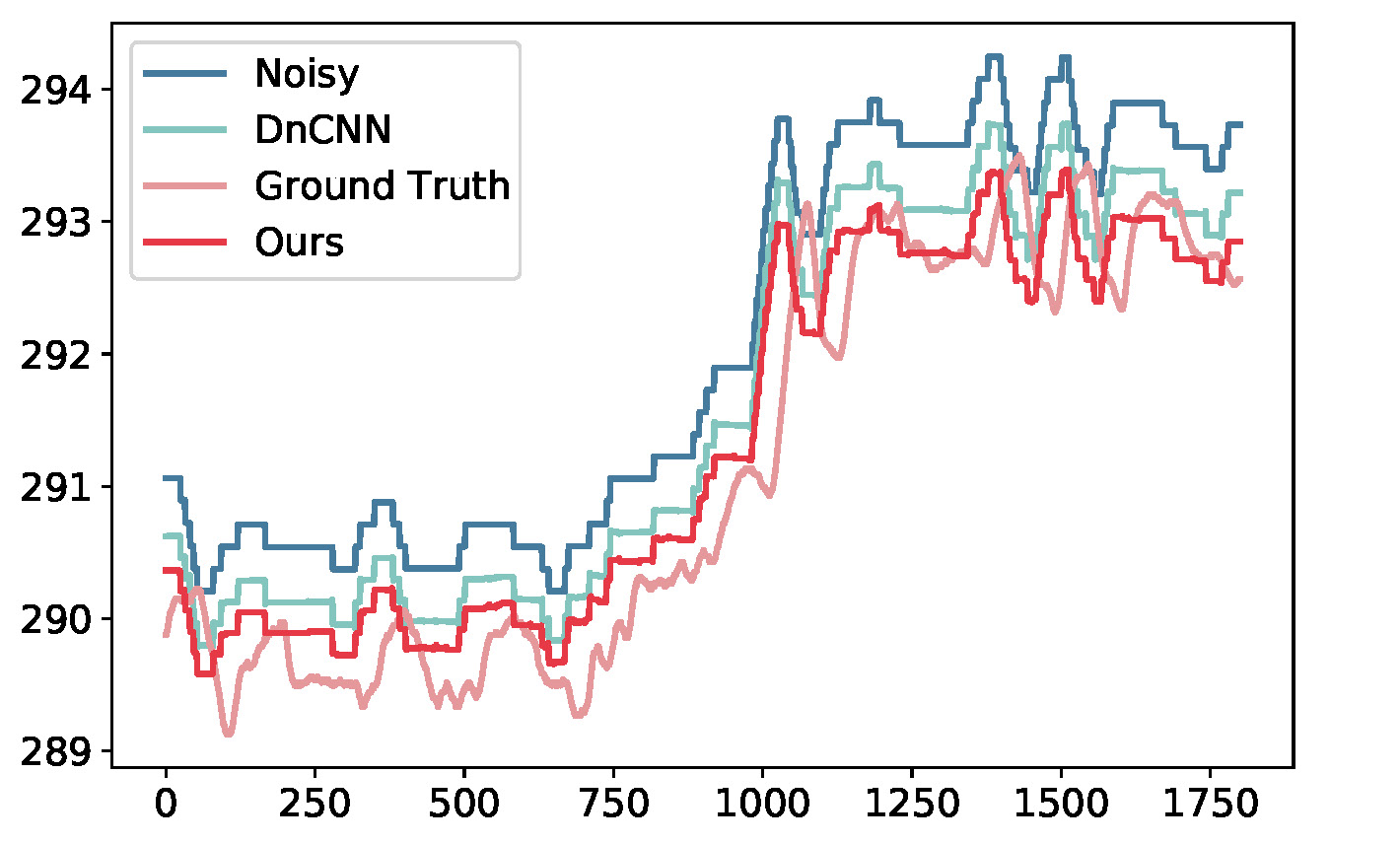}
          \caption{Example 1}
          \label{fig:hvac-t-sa_case1}
      \end{subfigure}
      \begin{subfigure}[b]{0.225\textwidth}
          \centering
          \includegraphics[width=\textwidth]{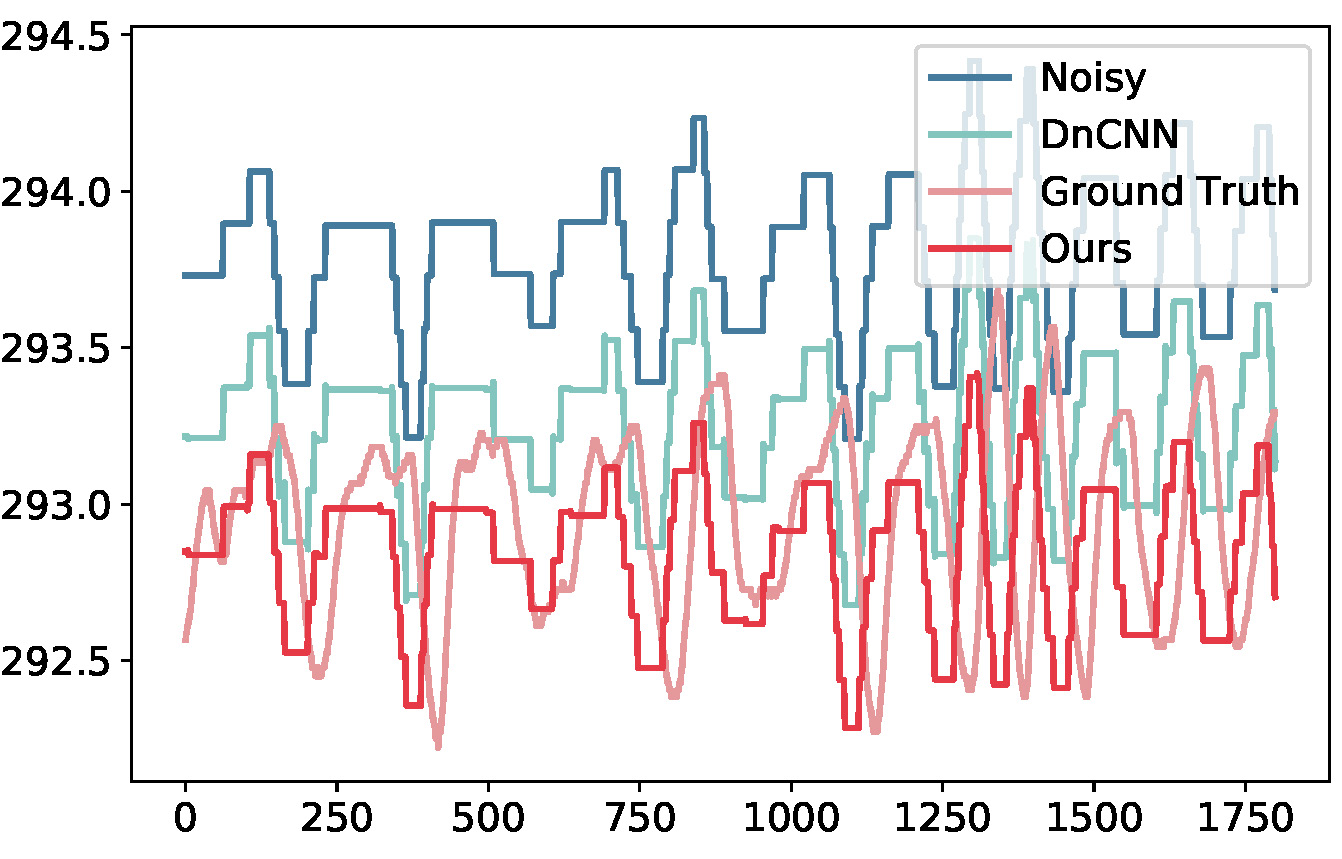}
          \caption{Example 2}
          \label{fig:hvac-t-sa_case2}
      \end{subfigure}
         \caption{Two example reconstructions for supply air temperature. \our’s outputs best align with the ground truth compared with original noisy measurements and the best- performing baseline. X-axis: timestep, Y-axis: K.} 
         \label{fig:hvac-t-sa}
      \centering
      \begin{subfigure}[b]{0.23\textwidth}
          \centering
          \includegraphics[width=\textwidth]{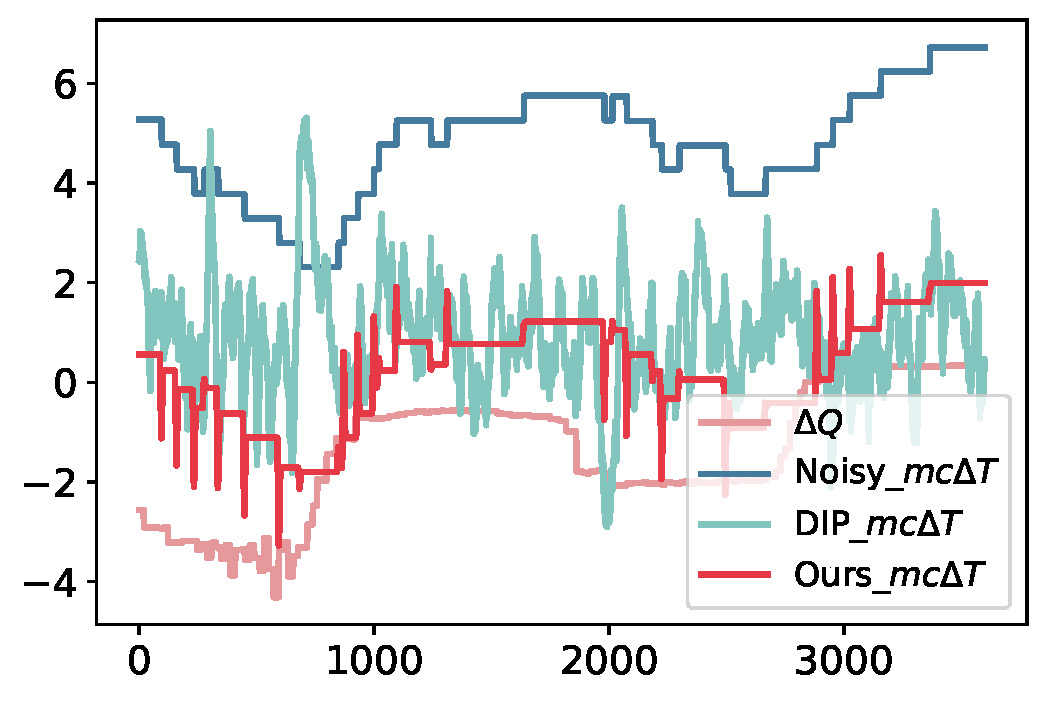}
          \caption{Example 1}
          \label{fig:hvac_case1}
      \end{subfigure}
      \begin{subfigure}[b]{0.23\textwidth}
          \centering
          \includegraphics[width=\textwidth]{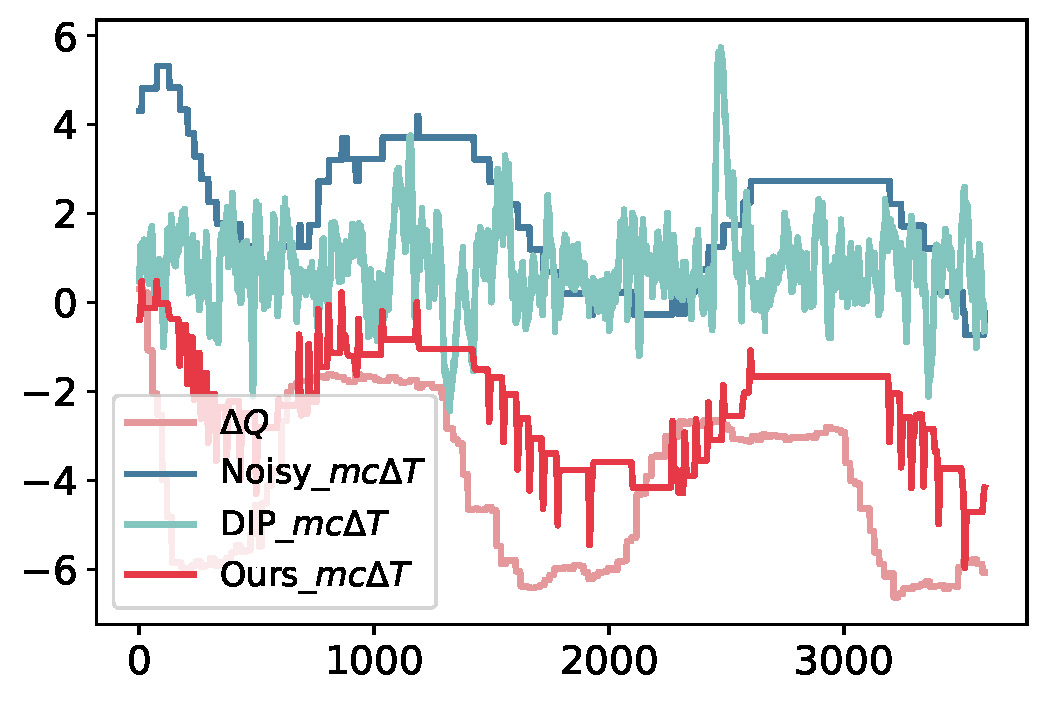}
          \caption{Example 2}
          \label{fig:hvac_case2}
      \end{subfigure}
         \caption{Two example alignment between $\Delta Q$ and $mc \Delta T$. \our achieves the closet alignment compared with original data and the best baseline. X-axis: timestep, Y-axis: K.} 
         \label{fig:hvac-case}
\end{figure}

\subsubsection{Reconstruction and Physics Performance}
For the HVAC control system application, we measure the denoising performance by both reconstruction performance and physics alignment. For reconstruction performance, we deploy high-precision temperature sensors and utilize the collected data as ground truth temperature to assess the reconstruction performance of different denoising methods. For physics alignment, we compute the MSE and MAE between $\Delta Q$ and $mc \Delta T$ in Equation~\ref{eq:hvac}. As shown in Table~\ref{tab:hvac}, \emph{\our achieves state-of-the-art performance and shows the lowest errors for both reconstruction performance and physics alignment}. We also provide qualitative comparisons to supplement our evaluation. Figure~\ref{fig:hvac-t-sa} showcases two example reconstructions of supply air temperature data. Denoised data from \our exhibits the best alignment with the ground truth data compared with original data, as well as the best-performing baseline DnCNN. In Figure~\ref{fig:hvac-case}, we also present two examples of physics alignment. We see that denoised data of \our best align with the physics equation compared with original noisy data and the best-performing baseline DIP.

\subsection{Ablation Study}

To study the effects of different components of our model, we separately remove the physics-based loss, reconstruction loss, and pre-training phase, and evaluate \our in Table~\ref{tab:abl}. For Inertial Navigation System (INS), we compare the physics alignment of both acceleration ($\mathrm{MSE_a}$, $\mathrm{MAE_a}$) and angular velocity ($\mathrm{MSE_w}$, $\mathrm{MAE_w}$). For CO$_2$ monitoring and HVAC control, we compare both reconstruction ($\mathrm{MSE_{rec}}$, $\mathrm{MAE_{rec}}$) and physics alignment ($\mathrm{MSE_{phy}}$, $\mathrm{MAE_{phy}}$). 

We note that by removing the physics-based loss, \our degenerates to a naive denoising model, which is not sufficient to capture the complex noise distribution in real-world sensor data. Therefore, compared with \our, MSE and MAE of both reconstruction and physics alignment increase after removing the physics-based loss. 
We also observed that a higher degree of precision in the physics model and higher frequency sensor data sampling (resulting in fewer synchronization errors), lead to larger performance improvements. For example, the relationships in inertial navigation physics are more precisely captured than with CO$_2$ or temperature systems, and the IMU sensors are sampled at higher frequencies than CO$_2$ sensors. Correspondingly, we observe greater performance gains from employing physics equations than other applications.
Secondly, reconstruction loss is important as it facilitates the model's ability to learn data distribution. In its absence, the model may potentially generate trivial outputs that, while satisfying physics constraints, neglect the actual data distribution (e.g., outputs of pure zeros). Lastly, the pre-training phase serves as an essential warm-up period to help the model better adapt to the underlying data distribution. To summarize, the physics-based loss, reconstruction loss, and pre-training phase collectively contribute to the overall efficacy of the model, and their presence is crucial for optimal performance.

\begin{table}[t]
\centering
    \caption{Ablation Study of \our. We bold the best and \underline{underline} the second best. Removing either physics-based loss, reconstruction loss or pre-training phase would negatively affect the performance, demonstrating the effectiveness of all three components.}
    \scalebox{0.9}{
    \setlength{\tabcolsep}{1.3mm}{
    \begin{tabular}{cccccc}
    \toprule
    Task & Metrics & w/o $l_{\mathrm{phy}}$ & w/o $l_{\mathrm{rec}}$ & w/o pre-train & \our \\
    \midrule
    \multirow{4}{*}{INS} & $\mathrm{MSE_a}$ & 316.5 & 259.1 & \underline{20.70} & \textbf{1.8695}\\
     & $\mathrm{MAE_a}$ & 8.342 & 8.759 & \underline{2.2930} & \textbf{0.6372}\\
     & $\mathrm{MSE_w}$ & 0.3579 & 0.2845 & \underline{0.1380} & \textbf{0.0380} \\
     & $\mathrm{MAE_w}$ & 0.1899 & 0.3352 & \underline{0.1291} & \textbf{0.0690}\\
     \midrule
     \multirow{4}{*}{CO$_2$} & $\mathrm{MSE_{rec}}$ &0.4641 & \underline{0.0568} &0.0724 &\textbf{0.0371}\\
     & $\mathrm{MAE_{rec}}$ & 0.0139 & \underline{0.0051} & 0.0074& \textbf{0.0047}\\
     & $\mathrm{MSE_{phy}}$ & 0.0194 & \underline{0.0021} &0.0023 &\textbf{0.0012}\\
     & $\mathrm{MAE_{phy}}$ & 0.0781 & \underline{0.0264} &0.0330 & \textbf{0.0200}\\
     \midrule
     \multirow{4}{*}{HVAC} &$\mathrm{MSE_{rec}}$ & 0.3314 & 0.5314  & \underline{0.2686} &\textbf{0.1994}\\
     & $\mathrm{MAE_{rec}}$ & 0.4709 & 0.6128 & \underline{0.4670} & \textbf{0.3454}\\
     & $\mathrm{MSE_{phy}}$ & 50.22 & 14.69 & \underline{14.469} & \textbf{14.081}\\
     & $\mathrm{MAE_{phy}}$ & 6.713 & 3.263 & \underline{3.203} & \textbf{3.1600}\\
    \bottomrule
    \end{tabular}}
    }
    \label{tab:abl}
\end{table}

\subsection{Sensitivity Analysis}\label{sec:sensitivity}

\begin{figure*}
    \centering
    \includegraphics[width=\linewidth]{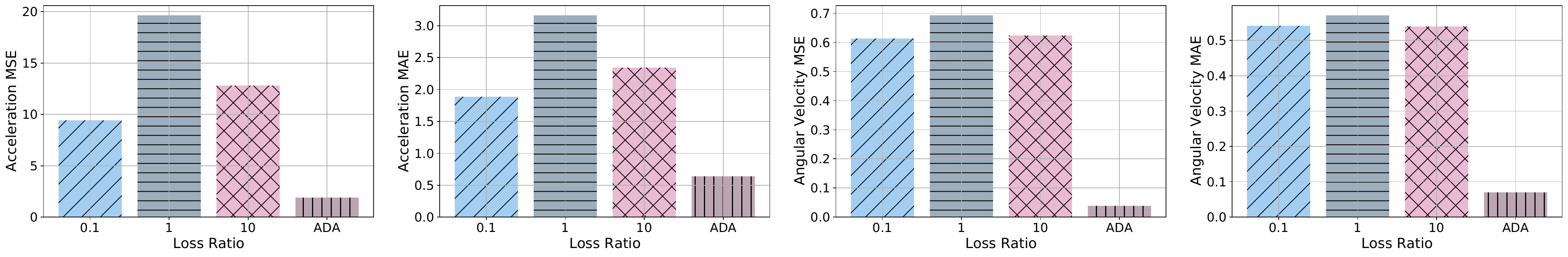}
    \caption{Performance analysis on the ratio of reconstruction loss over physics-based loss. Adaptive loss ratio yields better performance compared with fixed ratios. }
    \label{fig:loss}

    \centering
    \includegraphics[width=\linewidth]{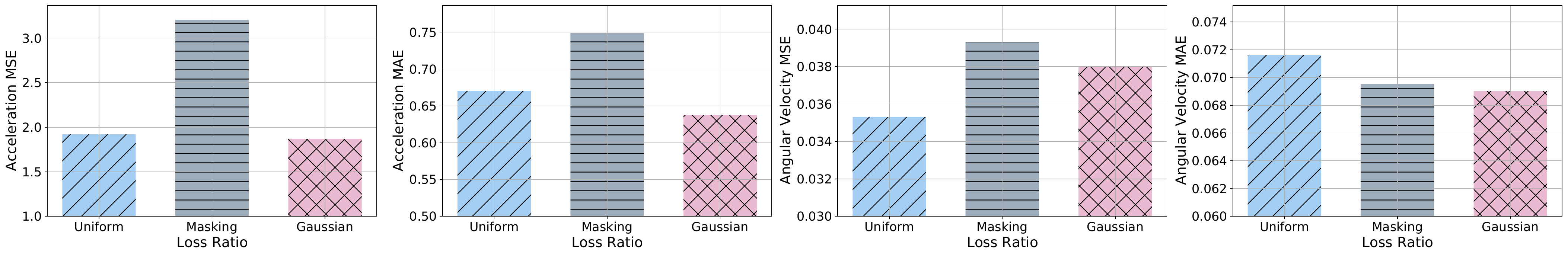}
    \caption{Performance analysis on the types of manually injected noise for the reconstruction process. Applying random noise performs better than zero masking.}
    \label{fig:noise}
\end{figure*}

\subsubsection{Balance between Reconstruction Loss and Physics-Based Loss}
In our experiments, we adopt an adaptive strategy to adjust the ratio between reconstruction loss ($l_{\mathrm{rec}}$) and physics-based loss ($l_{\mathrm{phy}}$) during training. Denote the ratio as $\lambda=l_{\mathrm{rec}}/l_{\mathrm{phy}}$. More specifically, for each iteration, we compute the mean value of the reconstruction loss, and adaptively adjust $\lambda$ such that ($\lambda \cdot l_{\mathrm{phy}}$) aligns with the same order of magnitude as $l_{\mathrm{rec}}$. Consequently, this adaptive approach avoids overemphasizing either the reconstruction loss or physics-based loss throughout the entire training process. To validate the effectiveness of adaptive loss ratios, we compare \our with its counterparts trained with fixed loss ratios, varying from 0.1, 1 to 10. As shown in Figure~\ref{fig:loss}, adaptive loss ratio (denoted as ``ADA'') yields the best performance compared with different fixed ratios. 

\subsubsection{Manually Injected Noise Types}
We also study the effect of different noise types injected for optimizing the reconstruction loss. In our experiments, we use Gaussian noise as an example injected noise. We compare \our with its counterparts trained with manually injected uniformly distributed noise, as well as manually injected zero masks. As shown in Figure~\ref{fig:noise}, injecting random noise (Gaussian or Uniform noise) performs better than applying zero masking, as random noise aligns closer to real-world noise distribution compared with zero masking. Moreover, the model is less sensitive to the particular noise types applied, such as Gaussian noise or uniformly distributed noise.

\subsection{Efficiency Analysis for Edge Devices}\label{sec:efficiency}

\begin{table}[t]
\centering
    \caption{Efficiency Analysis on Raspberry Pi 4.}
    \setlength{\tabcolsep}{1.1mm}{
    \begin{tabular}{ccccc}
    \toprule
    Metrics & Params & Size & Inference Time & CPU Usage \\
    \midrule
    Efficiency & 270K & 284 KB & 4 ms & 25\% \\
    \bottomrule
    \end{tabular}
    }
    \label{tab:complexity}
\end{table}

As many denoising methods for sensor data run on edge devices, we explore the time and memory efficiency of \our in Table~\ref{tab:complexity} using Raspberry Pi 4 as an example edge device. We deploy the inertial navigation denoising application through the Tensorflow Lite framework on Raspberry Pi 4. For time efficiency, \our can denoise a sequence of 100 readings (corresponding to a sequence of 1 second) within 4 milliseconds. To ensure the reliability of our evaluation, this experiment has been repeated a thousand times to obtain the average inference time. Therefore, \our is capable of performing sensor data denoising in real time. For memory efficiency, \our is a lightweight CNN model with just 270K parameters and 284 KB model size. Moreover, it demonstrates an average CPU utilization of only 25\% during the entire inference process. 
In summary, \our is both time and memory efficient for real-time operations on edge devices.
\section{Discussions and Future Work}
In this section, we discuss potential future directions of \our.

\smallskip\noindent \textbf{Broader Applicability to Other Sensing Systems} We acknowledge that the need for clearly formulated physics relationships is one limitation of our work. However, we note that apart from the three applications we have explored, in practice numerous sensing systems have underlying physics relationships, and we recognize this opportunity for extended exploration in future work. We discuss a few potential use cases as some examples for the model's wider applicability:

\begin{itemize}[nosep,leftmargin=*]
\item \noindent \textbf{Power System.} In a power system, the most classical Ohm's law describes the relationships between power, current and voltage. Electronic devices measuring current or voltage are often subject to environmental disruptions, such as temperature fluctuations, resulting in measurement noise. Utilizing physics priors like Ohm's Law, \our can offer a promising solution to mitigate these noise-related inaccuracies.
\item \noindent \textbf{Weather Monitoring.} Weather monitoring and forecasting involve calculations with respect to air pressure, temperature, wind speed, etc. These properties can be modeled as fluid dynamics problems, which are often characterized by the Navier-Stokes equations linking velocity and pressure. Environmental sensors responsible for capturing data like wind speed or pressure are susceptible to noise originating from environmental influences or sensor sensitivity. \our can leverage physics-based priors such as the Navier-Stokes equations to denoise these collected data, enhancing the accuracy of weather monitoring.
\item \noindent \textbf{Localization in Autonomous Systems.} In autonomous systems, we employ a combination of GPS, Lidar, Radar and IMU sensors to localize autonomous driving cars or drones. These sensor readings can be noisy due to atmospheric effects, multipath propagation, etc. \our is capable of applying the laws of motion (e.g., acceleration is the second derivative of location) to denoise location data collected from these multiple sources. 
\end{itemize} 

\smallskip \noindent \textbf{Uncertainty Quantification} Our future work also includes incorporating uncertainty quantification (UQ) into the denoising process. The integration of UQ aims to provide a holistic understanding of noise structure, delivering not just a point estimate for the denoised data but also the model's confidence in that estimate, avoiding overconfident and potentially inaccurate inferences. The integration may also help quantify the effects of denoising with respect to the precision of the model or synchronization errors. Uncertainty quantification can be achieved by adopting a probabilistic approach to extend the current model architecture to output not just a single denoised signal but a distribution over possible denoised signals. \smallskip

\noindent \textbf{FPGA Acceleration} We have implemented our denoising method on edge devices like Raspberry Pi and demonstrated its time and memory efficiency. To take this step further, we plan to explore hardware acceleration as Field-Programmable Gate Arrays (FPGA) co-processors to explore the possibility and cost of such edge inference. 
With FPGA's reprogrammable nature, we have the flexibility of prototyping and testing different configurations and parameters for the best possible performance and efficiency. Furthermore, FPGA allows us to seamlessly transition from prototype to subsequent production of Application-Specific Integrated Circuit (ASIC), which enable us to scale up our solution while reducing power consumption and production costs. 

\section{Conclusion}
We presented a physics-informed denoising method, \our, for real-life sensing systems. We build upon the insight that measurements from different sensors are intrinsically related by known underlying physics principles. This approach allows the model to harness these physics constraints as guidance during training to improve denoising process. This paves the way for a more practical denoising solution, especially given the frequent challenges associated with acquiring ground truth clean data in sensing systems or understanding underlying noise patterns. 
Extensive experiments show the efficacy of \our in removing sensor noise across three representative real-world sensing systems.
\our produces denoised results for low-cost sensors that align closely with high-precision and high-cost sensors, leading to a cost-effective denoising approach. 
\our is also lightweight and can enable real-time denoising on edge devices.

\begin{acks}
Our work is supported in part by ACE, one of the seven centers in JUMP 2.0, a Semiconductor Research Corporation (SRC) program sponsored by DARPA. Our work is also supported by Qualcomm Innovation Fellowship and is sponsored by NSF CAREER Award 2239440, NSF Proto-OKN Award 2333790, NIH Bridge2AI Center Program under award 1U54HG012510-01, Cisco-UCSD Sponsored Research Project, as well as generous gifts from Google, Adobe, and Teradata. Any opinions, findings, and conclusions or recommendations expressed herein are those of the authors and should not be interpreted as necessarily representing the views, either expressed or implied, of the U.S. Government. The U.S. Government is authorized to reproduce and distribute reprints for government purposes not withstanding any copyright annotation hereon.
\end{acks}

\bibliographystyle{ACM-Reference-Format}
\bibliography{sample-base}


\begin{thebibliography}{66}


\ifx \showCODEN    \undefined \def \showCODEN     #1{\unskip}     \fi
\ifx \showDOI      \undefined \def \showDOI       #1{#1}\fi
\ifx \showISBNx    \undefined \def \showISBNx     #1{\unskip}     \fi
\ifx \showISBNxiii \undefined \def \showISBNxiii  #1{\unskip}     \fi
\ifx \showISSN     \undefined \def \showISSN      #1{\unskip}     \fi
\ifx \showLCCN     \undefined \def \showLCCN      #1{\unskip}     \fi
\ifx \shownote     \undefined \def \shownote      #1{#1}          \fi
\ifx \showarticletitle \undefined \def \showarticletitle #1{#1}   \fi
\ifx \showURL      \undefined \def \showURL       {\relax}        \fi
\providecommand\bibfield[2]{#2}
\providecommand\bibinfo[2]{#2}
\providecommand\natexlab[1]{#1}
\providecommand\showeprint[2][]{arXiv:#2}

\bibitem[ams OSRAM~AG(2016)]%
        {CCS811}
\bibfield{author}{\bibinfo{person}{ams OSRAM~AG}.} \bibinfo{year}{2016}\natexlab{}.
\newblock \bibinfo{title}{CCS811 Datasheet}.
\newblock \bibinfo{howpublished}{\url{https://cdn.sparkfun.com/assets/learn_tutorials/1/4/3/CCS811_Datasheet-DS000459.pdf}}.
\newblock


\bibitem[Arief-Ang et~al\mbox{.}(2017)]%
        {arief2017hoc}
\bibfield{author}{\bibinfo{person}{Irvan~B Arief-Ang}, \bibinfo{person}{Flora~D Salim}, {and} \bibinfo{person}{Margaret Hamilton}.} \bibinfo{year}{2017}\natexlab{}.
\newblock \showarticletitle{Da-hoc: semi-supervised domain adaptation for room occupancy prediction using co2 sensor data}. In \bibinfo{booktitle}{\emph{Proceedings of the 4th ACM International Conference on Systems for Energy-Efficient Built Environments}}. \bibinfo{pages}{1--10}.
\newblock


\bibitem[Beltran and Cerpa(2014)]%
        {beltran2014optimal}
\bibfield{author}{\bibinfo{person}{Alex Beltran} {and} \bibinfo{person}{Alberto~E Cerpa}.} \bibinfo{year}{2014}\natexlab{}.
\newblock \showarticletitle{Optimal HVAC building control with occupancy prediction}. In \bibinfo{booktitle}{\emph{Proceedings of the 1st ACM conference on embedded systems for energy-efficient buildings}}. \bibinfo{pages}{168--171}.
\newblock


\bibitem[Cao et~al\mbox{.}(2022)]%
        {cao2022reducio}
\bibfield{author}{\bibinfo{person}{Zhiwei Cao}, \bibinfo{person}{Ruihang Wang}, \bibinfo{person}{Xin Zhou}, {and} \bibinfo{person}{Yonggang Wen}.} \bibinfo{year}{2022}\natexlab{}.
\newblock \showarticletitle{Reducio: model reduction for data center predictive digital twins via physics-guided machine learning}. In \bibinfo{booktitle}{\emph{Proceedings of the 9th ACM International Conference on Systems for Energy-Efficient Buildings, Cities, and Transportation}}. \bibinfo{pages}{1--10}.
\newblock


\bibitem[Chen et~al\mbox{.}(2018a)]%
        {chen2018ionet}
\bibfield{author}{\bibinfo{person}{Changhao Chen}, \bibinfo{person}{Xiaoxuan Lu}, \bibinfo{person}{Andrew Markham}, {and} \bibinfo{person}{Niki Trigoni}.} \bibinfo{year}{2018}\natexlab{a}.
\newblock \showarticletitle{Ionet: Learning to cure the curse of drift in inertial odometry}. In \bibinfo{booktitle}{\emph{Proceedings of the AAAI Conference on Artificial Intelligence}}, Vol.~\bibinfo{volume}{32}.
\newblock


\bibitem[Chen et~al\mbox{.}(2018c)]%
        {chen2018oxiod}
\bibfield{author}{\bibinfo{person}{Changhao Chen}, \bibinfo{person}{Peijun Zhao}, \bibinfo{person}{Chris~Xiaoxuan Lu}, \bibinfo{person}{Wei Wang}, \bibinfo{person}{Andrew Markham}, {and} \bibinfo{person}{Niki Trigoni}.} \bibinfo{year}{2018}\natexlab{c}.
\newblock \showarticletitle{Oxiod: The dataset for deep inertial odometry}.
\newblock \bibinfo{journal}{\emph{arXiv preprint arXiv:1809.07491}} (\bibinfo{year}{2018}).
\newblock


\bibitem[Chen et~al\mbox{.}(2020)]%
        {chen2020adaptive}
\bibfield{author}{\bibinfo{person}{Mingkang Chen}, \bibinfo{person}{Jingtao Sun}, \bibinfo{person}{Kazushige Saga}, \bibinfo{person}{Tomota Tanjo}, {and} \bibinfo{person}{Kento Aida}.} \bibinfo{year}{2020}\natexlab{}.
\newblock \showarticletitle{An adaptive noise removal tool for iot image processing under influence of weather conditions}. In \bibinfo{booktitle}{\emph{Proceedings of the 18th Conference on Embedded Networked Sensor Systems}}. \bibinfo{pages}{655--656}.
\newblock


\bibitem[Chen et~al\mbox{.}(2018b)]%
        {chen2018pga}
\bibfield{author}{\bibinfo{person}{Xinlei Chen}, \bibinfo{person}{Xiangxiang Xu}, \bibinfo{person}{Xinyu Liu}, \bibinfo{person}{Shijia Pan}, \bibinfo{person}{Jiayou He}, \bibinfo{person}{Hae~Young Noh}, \bibinfo{person}{Lin Zhang}, {and} \bibinfo{person}{Pei Zhang}.} \bibinfo{year}{2018}\natexlab{b}.
\newblock \showarticletitle{Pga: Physics guided and adaptive approach for mobile fine-grained air pollution estimation}. In \bibinfo{booktitle}{\emph{Proceedings of the 2018 ACM International Joint Conference and 2018 International Symposium on Pervasive and Ubiquitous Computing and Wearable Computers}}. \bibinfo{pages}{1321--1330}.
\newblock


\bibitem[Collier-Oxandale et~al\mbox{.}(2019a)]%
        {amt-12-1441-2019}
\bibfield{author}{\bibinfo{person}{A.~M. Collier-Oxandale}, \bibinfo{person}{J. Thorson}, \bibinfo{person}{H. Halliday}, \bibinfo{person}{J. Milford}, {and} \bibinfo{person}{M. Hannigan}.} \bibinfo{year}{2019}\natexlab{a}.
\newblock \showarticletitle{Understanding the ability of low-cost MOx sensors to quantify ambient VOCs}.
\newblock \bibinfo{journal}{\emph{Atmospheric Measurement Techniques}} \bibinfo{volume}{12}, \bibinfo{number}{3} (\bibinfo{year}{2019}), \bibinfo{pages}{1441--1460}.
\newblock
\urldef\tempurl%
\url{https://doi.org/10.5194/amt-12-1441-2019}
\showDOI{\tempurl}


\bibitem[Collier-Oxandale et~al\mbox{.}(2019b)]%
        {MOX}
\bibfield{author}{\bibinfo{person}{A.~M. Collier-Oxandale}, \bibinfo{person}{J. Thorson}, \bibinfo{person}{H. Halliday}, \bibinfo{person}{J. Milford}, {and} \bibinfo{person}{M. Hannigan}.} \bibinfo{year}{2019}\natexlab{b}.
\newblock \showarticletitle{Understanding the ability of low-cost MOx sensors to quantify ambient VOCs}.
\newblock \bibinfo{journal}{\emph{Atmospheric Measurement Techniques}} \bibinfo{volume}{12}, \bibinfo{number}{3} (\bibinfo{year}{2019}), \bibinfo{pages}{1441--1460}.
\newblock
\urldef\tempurl%
\url{https://doi.org/10.5194/amt-12-1441-2019}
\showDOI{\tempurl}


\bibitem[De~Faria et~al\mbox{.}(2017)]%
        {de2017insights}
\bibfield{author}{\bibinfo{person}{Maria Lu{\'\i}sa~Lopes De~Faria}, \bibinfo{person}{Carlos~Eduardo Cugnasca}, {and} \bibinfo{person}{Jos{\'e} Roberto~Almeida Amazonas}.} \bibinfo{year}{2017}\natexlab{}.
\newblock \showarticletitle{Insights into IoT data and an innovative DWT-based technique to denoise sensor signals}.
\newblock \bibinfo{journal}{\emph{IEEE Sensors Journal}} \bibinfo{volume}{18}, \bibinfo{number}{1} (\bibinfo{year}{2017}), \bibinfo{pages}{237--247}.
\newblock


\bibitem[Falcao et~al\mbox{.}(2021)]%
        {falcao2021piwims}
\bibfield{author}{\bibinfo{person}{Joao~Diogo Falcao}, \bibinfo{person}{Prabh Simran~S Baweja}, \bibinfo{person}{Yi Wang}, \bibinfo{person}{Akkarit Sangpetch}, \bibinfo{person}{Hae~Young Noh}, \bibinfo{person}{Orathai Sangpetch}, {and} \bibinfo{person}{Pei Zhang}.} \bibinfo{year}{2021}\natexlab{}.
\newblock \showarticletitle{PIWIMS: Physics Informed Warehouse Inventory Monitory via Synthetic Data Generation}. In \bibinfo{booktitle}{\emph{Adjunct Proceedings of the 2021 ACM International Joint Conference on Pervasive and Ubiquitous Computing and Proceedings of the 2021 ACM International Symposium on Wearable Computers}}. \bibinfo{pages}{613--618}.
\newblock


\bibitem[Faulkner et~al\mbox{.}(2013)]%
        {faulkner2013fresh}
\bibfield{author}{\bibinfo{person}{Matthew Faulkner}, \bibinfo{person}{Annie~H Liu}, {and} \bibinfo{person}{Andreas Krause}.} \bibinfo{year}{2013}\natexlab{}.
\newblock \showarticletitle{A fresh perspective: Learning to sparsify for detection in massive noisy sensor networks}. In \bibinfo{booktitle}{\emph{Proceedings of the 12th international conference on Information processing in sensor networks}}. \bibinfo{pages}{7--18}.
\newblock


\bibitem[Fu et~al\mbox{.}(2020)]%
        {10.1145/3408308.3431113}
\bibfield{author}{\bibinfo{person}{Xiaohan Fu}, \bibinfo{person}{Jason Koh}, \bibinfo{person}{Francesco Fraternali}, \bibinfo{person}{Dezhi Hong}, {and} \bibinfo{person}{Rajesh Gupta}.} \bibinfo{year}{2020}\natexlab{}.
\newblock \showarticletitle{Zonal Air Handling in Commercial Buildings}. In \bibinfo{booktitle}{\emph{Proceedings of the 7th ACM International Conference on Systems for Energy-Efficient Buildings, Cities, and Transportation}} (Virtual Event, Japan) \emph{(\bibinfo{series}{BuildSys '20})}. \bibinfo{publisher}{Association for Computing Machinery}, \bibinfo{address}{New York, NY, USA}, \bibinfo{pages}{302–303}.
\newblock
\showISBNx{9781450380614}
\urldef\tempurl%
\url{https://doi.org/10.1145/3408308.3431113}
\showDOI{\tempurl}


\bibitem[Gao et~al\mbox{.}(2021)]%
        {gao2021super}
\bibfield{author}{\bibinfo{person}{Han Gao}, \bibinfo{person}{Luning Sun}, {and} \bibinfo{person}{Jian-Xun Wang}.} \bibinfo{year}{2021}\natexlab{}.
\newblock \showarticletitle{Super-resolution and denoising of fluid flow using physics-informed convolutional neural networks without high-resolution labels}.
\newblock \bibinfo{journal}{\emph{Physics of Fluids}} \bibinfo{volume}{33}, \bibinfo{number}{7} (\bibinfo{year}{2021}), \bibinfo{pages}{073603}.
\newblock


\bibitem[Garcia~Satorras et~al\mbox{.}(2019)]%
        {garcia2019combining}
\bibfield{author}{\bibinfo{person}{Victor Garcia~Satorras}, \bibinfo{person}{Zeynep Akata}, {and} \bibinfo{person}{Max Welling}.} \bibinfo{year}{2019}\natexlab{}.
\newblock \showarticletitle{Combining generative and discriminative models for hybrid inference}.
\newblock \bibinfo{journal}{\emph{Advances in Neural Information Processing Systems}}  \bibinfo{volume}{32} (\bibinfo{year}{2019}).
\newblock


\bibitem[Hao et~al\mbox{.}(2022)]%
        {hao2022physics}
\bibfield{author}{\bibinfo{person}{Zhongkai Hao}, \bibinfo{person}{Songming Liu}, \bibinfo{person}{Yichi Zhang}, \bibinfo{person}{Chengyang Ying}, \bibinfo{person}{Yao Feng}, \bibinfo{person}{Hang Su}, {and} \bibinfo{person}{Jun Zhu}.} \bibinfo{year}{2022}\natexlab{}.
\newblock \showarticletitle{Physics-Informed Machine Learning: A Survey on Problems, Methods and Applications}.
\newblock \bibinfo{journal}{\emph{arXiv preprint arXiv:2211.08064}} (\bibinfo{year}{2022}).
\newblock


\bibitem[He et~al\mbox{.}(2020)]%
        {he2020scsv2}
\bibfield{author}{\bibinfo{person}{Lixing He}, \bibinfo{person}{Carlos Ruiz}, \bibinfo{person}{Mostafa Mirshekari}, {and} \bibinfo{person}{Shijia Pan}.} \bibinfo{year}{2020}\natexlab{}.
\newblock \showarticletitle{Scsv2: physics-informed self-configuration sensing through vision and vibration context modeling}. In \bibinfo{booktitle}{\emph{Adjunct Proceedings of the 2020 ACM International Joint Conference on Pervasive and Ubiquitous Computing and Proceedings of the 2020 ACM International Symposium on Wearable Computers}}. \bibinfo{pages}{532--537}.
\newblock


\bibitem[Herath et~al\mbox{.}(2020)]%
        {herath2020ronin}
\bibfield{author}{\bibinfo{person}{Sachini Herath}, \bibinfo{person}{Hang Yan}, {and} \bibinfo{person}{Yasutaka Furukawa}.} \bibinfo{year}{2020}\natexlab{}.
\newblock \showarticletitle{Ronin: Robust neural inertial navigation in the wild: Benchmark, evaluations, \& new methods}. In \bibinfo{booktitle}{\emph{2020 IEEE International Conference on Robotics and Automation (ICRA)}}. IEEE, \bibinfo{pages}{3146--3152}.
\newblock


\bibitem[Huang et~al\mbox{.}(2021)]%
        {huang2021neighbor2neighbor}
\bibfield{author}{\bibinfo{person}{Tao Huang}, \bibinfo{person}{Songjiang Li}, \bibinfo{person}{Xu Jia}, \bibinfo{person}{Huchuan Lu}, {and} \bibinfo{person}{Jianzhuang Liu}.} \bibinfo{year}{2021}\natexlab{}.
\newblock \showarticletitle{Neighbor2neighbor: Self-supervised denoising from single noisy images}. In \bibinfo{booktitle}{\emph{Proceedings of the IEEE/CVF conference on computer vision and pattern recognition}}. \bibinfo{pages}{14781--14790}.
\newblock


\bibitem[Jeyakumar et~al\mbox{.}(2019)]%
        {jeyakumar2019sensehar}
\bibfield{author}{\bibinfo{person}{Jeya~Vikranth Jeyakumar}, \bibinfo{person}{Liangzhen Lai}, \bibinfo{person}{Naveen Suda}, {and} \bibinfo{person}{Mani Srivastava}.} \bibinfo{year}{2019}\natexlab{}.
\newblock \showarticletitle{SenseHAR: a robust virtual activity sensor for smartphones and wearables}. In \bibinfo{booktitle}{\emph{Proceedings of the 17th Conference on Embedded Networked Sensor Systems}}. \bibinfo{pages}{15--28}.
\newblock


\bibitem[Karniadakis et~al\mbox{.}(2021)]%
        {karniadakis2021physics}
\bibfield{author}{\bibinfo{person}{GE Karniadakis}, \bibinfo{person}{IG Kevrekidis}, \bibinfo{person}{L Lu}, \bibinfo{person}{P Perdikaris}, \bibinfo{person}{S Wang}, {and} \bibinfo{person}{L Yang}.} \bibinfo{year}{2021}\natexlab{}.
\newblock \showarticletitle{Physics-informed machine learning: Nature Reviews Physics}.
\newblock  (\bibinfo{year}{2021}).
\newblock


\bibitem[Kelshaw and Magri(2022)]%
        {kelshaw2022physics}
\bibfield{author}{\bibinfo{person}{Daniel Kelshaw} {and} \bibinfo{person}{Luca Magri}.} \bibinfo{year}{2022}\natexlab{}.
\newblock \showarticletitle{Physics-Informed Convolutional Neural Networks for Corruption Removal on Dynamical Systems}.
\newblock \bibinfo{journal}{\emph{arXiv preprint arXiv:2210.16215}} (\bibinfo{year}{2022}).
\newblock


\bibitem[Kim et~al\mbox{.}(2021)]%
        {kim2021physics}
\bibfield{author}{\bibinfo{person}{Minsung Kim}, \bibinfo{person}{Salvatore Mandr{\`a}}, \bibinfo{person}{Davide Venturelli}, {and} \bibinfo{person}{Kyle Jamieson}.} \bibinfo{year}{2021}\natexlab{}.
\newblock \showarticletitle{Physics-inspired heuristics for soft MIMO detection in 5G new radio and beyond}. In \bibinfo{booktitle}{\emph{Proceedings of the 27th Annual International Conference on Mobile Computing and Networking}}. \bibinfo{pages}{42--55}.
\newblock


\bibitem[Kurte et~al\mbox{.}(2021)]%
        {kurte2021comparative}
\bibfield{author}{\bibinfo{person}{Kuldeep Kurte}, \bibinfo{person}{Kadir Amasyali}, \bibinfo{person}{Jeffrey Munk}, {and} \bibinfo{person}{Helia Zandi}.} \bibinfo{year}{2021}\natexlab{}.
\newblock \showarticletitle{Comparative analysis of model-free and model-based HVAC control for residential demand response}. In \bibinfo{booktitle}{\emph{Proceedings of the 8th ACM international conference on systems for energy-efficient buildings, cities, and transportation}}. \bibinfo{pages}{309--313}.
\newblock


\bibitem[Lehtinen et~al\mbox{.}(2018)]%
        {lehtinen2018noise2noise}
\bibfield{author}{\bibinfo{person}{Jaakko Lehtinen}, \bibinfo{person}{Jacob Munkberg}, \bibinfo{person}{Jon Hasselgren}, \bibinfo{person}{Samuli Laine}, \bibinfo{person}{Tero Karras}, \bibinfo{person}{Miika Aittala}, {and} \bibinfo{person}{Timo Aila}.} \bibinfo{year}{2018}\natexlab{}.
\newblock \showarticletitle{Noise2Noise: Learning image restoration without clean data}.
\newblock \bibinfo{journal}{\emph{arXiv preprint arXiv:1803.04189}} (\bibinfo{year}{2018}).
\newblock


\bibitem[Li et~al\mbox{.}(2022)]%
        {li2022sqee}
\bibfield{author}{\bibinfo{person}{Shuheng Li}, \bibinfo{person}{Jingbo Shang}, \bibinfo{person}{Rajesh~K Gupta}, {and} \bibinfo{person}{Dezhi Hong}.} \bibinfo{year}{2022}\natexlab{}.
\newblock \showarticletitle{SQEE: A Machine Perception Approach to Sensing Quality Evaluation at the Edge by Uncertainty Quantification}. In \bibinfo{booktitle}{\emph{Proceedings of the 20th ACM Conference on Embedded Networked Sensor Systems}}. \bibinfo{pages}{277--290}.
\newblock


\bibitem[Liu et~al\mbox{.}(2022)]%
        {liu2022real}
\bibfield{author}{\bibinfo{person}{Miaomiao Liu}, \bibinfo{person}{Sikai Yang}, \bibinfo{person}{Wyssanie Chomsin}, {and} \bibinfo{person}{Wan Du}.} \bibinfo{year}{2022}\natexlab{}.
\newblock \showarticletitle{Real-time tracking of smartwatch orientation and location by multitask learning}. In \bibinfo{booktitle}{\emph{Proceedings of the 20th ACM Conference on Embedded Networked Sensor Systems}}. \bibinfo{pages}{120--133}.
\newblock


\bibitem[Liu et~al\mbox{.}(2021)]%
        {liu2021wavoice}
\bibfield{author}{\bibinfo{person}{Tiantian Liu}, \bibinfo{person}{Ming Gao}, \bibinfo{person}{Feng Lin}, \bibinfo{person}{Chao Wang}, \bibinfo{person}{Zhongjie Ba}, \bibinfo{person}{Jinsong Han}, \bibinfo{person}{Wenyao Xu}, {and} \bibinfo{person}{Kui Ren}.} \bibinfo{year}{2021}\natexlab{}.
\newblock \showarticletitle{Wavoice: A noise-resistant multi-modal speech recognition system fusing mmwave and audio signals}. In \bibinfo{booktitle}{\emph{Proceedings of the 19th ACM Conference on Embedded Networked Sensor Systems}}. \bibinfo{pages}{97--110}.
\newblock


\bibitem[Liu et~al\mbox{.}(2020)]%
        {liu2020tlio}
\bibfield{author}{\bibinfo{person}{Wenxin Liu}, \bibinfo{person}{David Caruso}, \bibinfo{person}{Eddy Ilg}, \bibinfo{person}{Jing Dong}, \bibinfo{person}{Anastasios~I Mourikis}, \bibinfo{person}{Kostas Daniilidis}, \bibinfo{person}{Vijay Kumar}, {and} \bibinfo{person}{Jakob Engel}.} \bibinfo{year}{2020}\natexlab{}.
\newblock \showarticletitle{Tlio: Tight learned inertial odometry}.
\newblock \bibinfo{journal}{\emph{IEEE Robotics and Automation Letters}} \bibinfo{volume}{5}, \bibinfo{number}{4} (\bibinfo{year}{2020}), \bibinfo{pages}{5653--5660}.
\newblock


\bibitem[Liu et~al\mbox{.}(2019)]%
        {liu2019real}
\bibfield{author}{\bibinfo{person}{Yang Liu}, \bibinfo{person}{Zhenjiang Li}, \bibinfo{person}{Zhidan Liu}, {and} \bibinfo{person}{Kaishun Wu}.} \bibinfo{year}{2019}\natexlab{}.
\newblock \showarticletitle{Real-time arm skeleton tracking and gesture inference tolerant to missing wearable sensors}. In \bibinfo{booktitle}{\emph{Proceedings of the 17th Annual International Conference on Mobile Systems, Applications, and Services}}. \bibinfo{pages}{287--299}.
\newblock


\bibitem[Lu et~al\mbox{.}(2020)]%
        {lu2020milliego}
\bibfield{author}{\bibinfo{person}{Chris~Xiaoxuan Lu}, \bibinfo{person}{Muhamad Risqi~U Saputra}, \bibinfo{person}{Peijun Zhao}, \bibinfo{person}{Yasin Almalioglu}, \bibinfo{person}{Pedro~PB De~Gusmao}, \bibinfo{person}{Changhao Chen}, \bibinfo{person}{Ke Sun}, \bibinfo{person}{Niki Trigoni}, {and} \bibinfo{person}{Andrew Markham}.} \bibinfo{year}{2020}\natexlab{}.
\newblock \showarticletitle{milliEgo: single-chip mmWave radar aided egomotion estimation via deep sensor fusion}. In \bibinfo{booktitle}{\emph{Proceedings of the 18th Conference on Embedded Networked Sensor Systems}}. \bibinfo{pages}{109--122}.
\newblock


\bibitem[Luo et~al\mbox{.}(2021)]%
        {luo2021phyaug}
\bibfield{author}{\bibinfo{person}{Wenjie Luo}, \bibinfo{person}{Zhenyu Yan}, \bibinfo{person}{Qun Song}, {and} \bibinfo{person}{Rui Tan}.} \bibinfo{year}{2021}\natexlab{}.
\newblock \showarticletitle{Phyaug: Physics-directed data augmentation for deep sensing model transfer in cyber-physical systems}. In \bibinfo{booktitle}{\emph{Proceedings of the 20th International Conference on Information Processing in Sensor Networks (co-located with CPS-IoT Week 2021)}}. \bibinfo{pages}{31--46}.
\newblock


\bibitem[Moran et~al\mbox{.}(2020)]%
        {moran2020noisier2noise}
\bibfield{author}{\bibinfo{person}{Nick Moran}, \bibinfo{person}{Dan Schmidt}, \bibinfo{person}{Yu Zhong}, {and} \bibinfo{person}{Patrick Coady}.} \bibinfo{year}{2020}\natexlab{}.
\newblock \showarticletitle{Noisier2noise: Learning to denoise from unpaired noisy data}. In \bibinfo{booktitle}{\emph{Proceedings of the IEEE/CVF Conference on Computer Vision and Pattern Recognition}}. \bibinfo{pages}{12064--12072}.
\newblock


\bibitem[Nagarathinam et~al\mbox{.}(2022)]%
        {nagarathinam2022pacman}
\bibfield{author}{\bibinfo{person}{Srinarayana Nagarathinam}, \bibinfo{person}{Yashovardhan~S Chati}, \bibinfo{person}{Malini~Pooni Venkat}, {and} \bibinfo{person}{Arunchandar Vasan}.} \bibinfo{year}{2022}\natexlab{}.
\newblock \showarticletitle{PACMAN: physics-aware control MANager for HVAC}. In \bibinfo{booktitle}{\emph{Proceedings of the 9th ACM International Conference on Systems for Energy-Efficient Buildings, Cities, and Transportation}}. \bibinfo{pages}{11--20}.
\newblock


\bibitem[Nagarathinam et~al\mbox{.}(2015)]%
        {nagarathinam2015centralized}
\bibfield{author}{\bibinfo{person}{Srinarayana Nagarathinam}, \bibinfo{person}{Arunchandar Vasan}, \bibinfo{person}{Venkata Ramakrishna~P}, \bibinfo{person}{Shiva~R Iyer}, \bibinfo{person}{Venkatesh Sarangan}, {and} \bibinfo{person}{Anand Sivasubramaniam}.} \bibinfo{year}{2015}\natexlab{}.
\newblock \showarticletitle{Centralized management of HVAC energy in large multi-AHU zones}. In \bibinfo{booktitle}{\emph{Proceedings of the 2nd ACM International Conference on Embedded Systems for Energy-Efficient Built Environments}}. \bibinfo{pages}{157--166}.
\newblock


\bibitem[of~Health~Service(2023)]%
        {co2_airquality}
\bibfield{author}{\bibinfo{person}{Wisconsin~Department of Health~Service}.} \bibinfo{year}{2023}\natexlab{}.
\newblock \bibinfo{title}{Carbon Dioxide}.
\newblock \bibinfo{howpublished}{\url{https://www.dhs.wisconsin.gov/chemical/carbondioxide.htm}}.
\newblock


\bibitem[Omitaomu et~al\mbox{.}(2011)]%
        {omitaomu2011empirical}
\bibfield{author}{\bibinfo{person}{Olufemi~A Omitaomu}, \bibinfo{person}{Vladimir~A Protopopescu}, {and} \bibinfo{person}{Auroop~R Ganguly}.} \bibinfo{year}{2011}\natexlab{}.
\newblock \showarticletitle{Empirical mode decomposition technique with conditional mutual information for denoising operational sensor data}.
\newblock \bibinfo{journal}{\emph{IEEE sensors journal}} \bibinfo{volume}{11}, \bibinfo{number}{10} (\bibinfo{year}{2011}), \bibinfo{pages}{2565--2575}.
\newblock


\bibitem[Palzer(2020)]%
        {Palzer2020-su}
\bibfield{author}{\bibinfo{person}{Stefan Palzer}.} \bibinfo{year}{2020}\natexlab{}.
\newblock \showarticletitle{Photoacoustic-based gas sensing: A review}.
\newblock \bibinfo{journal}{\emph{Sensors (Basel)}} \bibinfo{volume}{20}, \bibinfo{number}{9} (\bibinfo{date}{May} \bibinfo{year}{2020}), \bibinfo{pages}{2745}.
\newblock


\bibitem[Park et~al\mbox{.}(2017)]%
        {park2017glasses}
\bibfield{author}{\bibinfo{person}{Jaeyeon Park}, \bibinfo{person}{Woojin Nam}, \bibinfo{person}{Jaewon Choi}, \bibinfo{person}{Taeyeong Kim}, \bibinfo{person}{Dukyong Yoon}, \bibinfo{person}{Sukhoon Lee}, \bibinfo{person}{Jeongyeup Paek}, {and} \bibinfo{person}{JeongGil Ko}.} \bibinfo{year}{2017}\natexlab{}.
\newblock \showarticletitle{Glasses for the third eye: Improving the quality of clinical data analysis with motion sensor-based data filtering}. In \bibinfo{booktitle}{\emph{Proceedings of the 15th ACM Conference on Embedded Network Sensor Systems}}. \bibinfo{pages}{1--14}.
\newblock


\bibitem[Raissi et~al\mbox{.}(2019)]%
        {raissi2019physics}
\bibfield{author}{\bibinfo{person}{Maziar Raissi}, \bibinfo{person}{Paris Perdikaris}, {and} \bibinfo{person}{George~E Karniadakis}.} \bibinfo{year}{2019}\natexlab{}.
\newblock \showarticletitle{Physics-informed neural networks: A deep learning framework for solving forward and inverse problems involving nonlinear partial differential equations}.
\newblock \bibinfo{journal}{\emph{Journal of Computational physics}}  \bibinfo{volume}{378} (\bibinfo{year}{2019}), \bibinfo{pages}{686--707}.
\newblock


\bibitem[Ren et~al\mbox{.}(2018)]%
        {ren2018learning}
\bibfield{author}{\bibinfo{person}{Hongyu Ren}, \bibinfo{person}{Russell Stewart}, \bibinfo{person}{Jiaming Song}, \bibinfo{person}{Volodymyr Kuleshov}, {and} \bibinfo{person}{Stefano Ermon}.} \bibinfo{year}{2018}\natexlab{}.
\newblock \showarticletitle{Learning with weak supervision from physics and data-driven constraints}.
\newblock \bibinfo{journal}{\emph{AI Magazine}} \bibinfo{volume}{39}, \bibinfo{number}{1} (\bibinfo{year}{2018}), \bibinfo{pages}{27--38}.
\newblock


\bibitem[Rudnick and Milton(2003)]%
        {Rudnick2003-uo}
\bibfield{author}{\bibinfo{person}{S~N Rudnick} {and} \bibinfo{person}{D~K Milton}.} \bibinfo{year}{2003}\natexlab{}.
\newblock \showarticletitle{Risk of indoor airborne infection transmission estimated from carbon dioxide concentration}.
\newblock \bibinfo{journal}{\emph{Indoor Air}} \bibinfo{volume}{13}, \bibinfo{number}{3} (\bibinfo{date}{Sept.} \bibinfo{year}{2003}), \bibinfo{pages}{237--245}.
\newblock


\bibitem[Shen et~al\mbox{.}(2018)]%
        {shen2018closing}
\bibfield{author}{\bibinfo{person}{Sheng Shen}, \bibinfo{person}{Mahanth Gowda}, {and} \bibinfo{person}{Romit Roy~Choudhury}.} \bibinfo{year}{2018}\natexlab{}.
\newblock \showarticletitle{Closing the gaps in inertial motion tracking}. In \bibinfo{booktitle}{\emph{Proceedings of the 24th Annual International Conference on Mobile Computing and Networking}}. \bibinfo{pages}{429--444}.
\newblock


\bibitem[Shen et~al\mbox{.}(2016)]%
        {shen2016smartwatch}
\bibfield{author}{\bibinfo{person}{Sheng Shen}, \bibinfo{person}{He Wang}, {and} \bibinfo{person}{Romit Roy~Choudhury}.} \bibinfo{year}{2016}\natexlab{}.
\newblock \showarticletitle{I am a smartwatch and i can track my user's arm}. In \bibinfo{booktitle}{\emph{Proceedings of the 14th annual international conference on Mobile systems, applications, and services}}. \bibinfo{pages}{85--96}.
\newblock


\bibitem[Sun et~al\mbox{.}(2021)]%
        {sun2021idol}
\bibfield{author}{\bibinfo{person}{Scott Sun}, \bibinfo{person}{Dennis Melamed}, {and} \bibinfo{person}{Kris Kitani}.} \bibinfo{year}{2021}\natexlab{}.
\newblock \showarticletitle{Idol: Inertial deep orientation-estimation and localization}. In \bibinfo{booktitle}{\emph{Proceedings of the AAAI Conference on Artificial Intelligence}}, Vol.~\bibinfo{volume}{35}. \bibinfo{pages}{6128--6137}.
\newblock


\bibitem[Sun et~al\mbox{.}(2022)]%
        {sun2022c}
\bibfield{author}{\bibinfo{person}{Yifei Sun}, \bibinfo{person}{Yuxuan Liu}, \bibinfo{person}{Ziteng Wang}, \bibinfo{person}{Xiaolei Qu}, \bibinfo{person}{Dezhi Zheng}, {and} \bibinfo{person}{Xinlei Chen}.} \bibinfo{year}{2022}\natexlab{}.
\newblock \showarticletitle{C-RIDGE: Indoor CO2 Data Collection System for Large Venues Based on prior Knowledge}. In \bibinfo{booktitle}{\emph{Proceedings of the 20th ACM Conference on Embedded Networked Sensor Systems}}. \bibinfo{pages}{1077--1082}.
\newblock


\bibitem[Takeishi and Kalousis(2021)]%
        {takeishi2021physics}
\bibfield{author}{\bibinfo{person}{Naoya Takeishi} {and} \bibinfo{person}{Alexandros Kalousis}.} \bibinfo{year}{2021}\natexlab{}.
\newblock \showarticletitle{Physics-integrated variational autoencoders for robust and interpretable generative modeling}.
\newblock \bibinfo{journal}{\emph{Advances in Neural Information Processing Systems}}  \bibinfo{volume}{34} (\bibinfo{year}{2021}), \bibinfo{pages}{14809--14821}.
\newblock


\bibitem[Ulyanov et~al\mbox{.}(2018)]%
        {ulyanov2018deep}
\bibfield{author}{\bibinfo{person}{Dmitry Ulyanov}, \bibinfo{person}{Andrea Vedaldi}, {and} \bibinfo{person}{Victor Lempitsky}.} \bibinfo{year}{2018}\natexlab{}.
\newblock \showarticletitle{Deep image prior}. In \bibinfo{booktitle}{\emph{Proceedings of the IEEE conference on computer vision and pattern recognition}}. \bibinfo{pages}{9446--9454}.
\newblock


\bibitem[\url{CO2Meter.com}(2022)]%
        {K30}
\bibfield{author}{\bibinfo{person}{\url{CO2Meter.com}}.} \bibinfo{year}{2022}\natexlab{}.
\newblock \bibinfo{title}{K30 Datasheet}.
\newblock \bibinfo{howpublished}{\url{https://cdn.shopify.com/s/files/1/0019/5952/files/DS-SenseAir-K30-CO2-Sensor-Revised-061722_96d4f229-643c-45b0-95a6-24efa252490d.pdf?v=1672160950}}.
\newblock


\bibitem[Wang et~al\mbox{.}(2021)]%
        {wang2021tstnn}
\bibfield{author}{\bibinfo{person}{Kai Wang}, \bibinfo{person}{Bengbeng He}, {and} \bibinfo{person}{Wei-Ping Zhu}.} \bibinfo{year}{2021}\natexlab{}.
\newblock \showarticletitle{TSTNN: Two-stage transformer based neural network for speech enhancement in the time domain}. In \bibinfo{booktitle}{\emph{ICASSP 2021-2021 IEEE International Conference on Acoustics, Speech and Signal Processing (ICASSP)}}. IEEE, \bibinfo{pages}{7098--7102}.
\newblock


\bibitem[Wang et~al\mbox{.}(2020)]%
        {wang2020towards}
\bibfield{author}{\bibinfo{person}{Rui Wang}, \bibinfo{person}{Karthik Kashinath}, \bibinfo{person}{Mustafa Mustafa}, \bibinfo{person}{Adrian Albert}, {and} \bibinfo{person}{Rose Yu}.} \bibinfo{year}{2020}\natexlab{}.
\newblock \showarticletitle{Towards physics-informed deep learning for turbulent flow prediction}. In \bibinfo{booktitle}{\emph{Proceedings of the 26th ACM SIGKDD International Conference on Knowledge Discovery \& Data Mining}}. \bibinfo{pages}{1457--1466}.
\newblock


\bibitem[Wang and Yu(2021)]%
        {wang2021physics}
\bibfield{author}{\bibinfo{person}{Rui Wang} {and} \bibinfo{person}{Rose Yu}.} \bibinfo{year}{2021}\natexlab{}.
\newblock \showarticletitle{Physics-guided deep learning for dynamical systems: A survey}.
\newblock \bibinfo{journal}{\emph{arXiv preprint arXiv:2107.01272}} (\bibinfo{year}{2021}).
\newblock


\bibitem[Wang et~al\mbox{.}(2022)]%
        {wang2022capricorn}
\bibfield{author}{\bibinfo{person}{Ziqi Wang}, \bibinfo{person}{Ankur Sarker}, \bibinfo{person}{Jason Wu}, \bibinfo{person}{Derek Hua}, \bibinfo{person}{Gaofeng Dong}, \bibinfo{person}{Akash~Deep Singh}, {and} \bibinfo{person}{Mani Srivastava}.} \bibinfo{year}{2022}\natexlab{}.
\newblock \showarticletitle{Capricorn: Towards Real-time Rich Scene Analysis Using RF-Vision Sensor Fusion}. In \bibinfo{booktitle}{\emph{Proceedings of the 20th ACM Conference on Embedded Networked Sensor Systems}}. \bibinfo{pages}{334--348}.
\newblock


\bibitem[Weber et~al\mbox{.}(2020)]%
        {weber2020detecting}
\bibfield{author}{\bibinfo{person}{Manuel Weber}, \bibinfo{person}{Christoph Doblander}, {and} \bibinfo{person}{Peter Mandl}.} \bibinfo{year}{2020}\natexlab{}.
\newblock \showarticletitle{Detecting building occupancy with synthetic environmental data}. In \bibinfo{booktitle}{\emph{Proceedings of the 7th ACM International Conference on Systems for Energy-Efficient Buildings, Cities, and Transportation}}. \bibinfo{pages}{324--325}.
\newblock


\bibitem[Weekly et~al\mbox{.}(2015)]%
        {weekly2015modeling}
\bibfield{author}{\bibinfo{person}{Kevin Weekly}, \bibinfo{person}{Nikolaos Bekiaris-Liberis}, \bibinfo{person}{Ming Jin}, {and} \bibinfo{person}{Alexandre~M Bayen}.} \bibinfo{year}{2015}\natexlab{}.
\newblock \showarticletitle{Modeling and estimation of the humans' effect on the CO 2 dynamics inside a conference room}.
\newblock \bibinfo{journal}{\emph{IEEE Transactions on Control Systems Technology}} \bibinfo{volume}{23}, \bibinfo{number}{5} (\bibinfo{year}{2015}), \bibinfo{pages}{1770--1781}.
\newblock


\bibitem[Xu et~al\mbox{.}(2020)]%
        {xu2020noisy}
\bibfield{author}{\bibinfo{person}{Jun Xu}, \bibinfo{person}{Yuan Huang}, \bibinfo{person}{Ming-Ming Cheng}, \bibinfo{person}{Li Liu}, \bibinfo{person}{Fan Zhu}, \bibinfo{person}{Zhou Xu}, {and} \bibinfo{person}{Ling Shao}.} \bibinfo{year}{2020}\natexlab{}.
\newblock \showarticletitle{Noisy-as-clean: Learning self-supervised denoising from corrupted image}.
\newblock \bibinfo{journal}{\emph{IEEE Transactions on Image Processing}}  \bibinfo{volume}{29} (\bibinfo{year}{2020}), \bibinfo{pages}{9316--9329}.
\newblock


\bibitem[Xu et~al\mbox{.}(2022)]%
        {xu2022accelerate}
\bibfield{author}{\bibinfo{person}{Shichao Xu}, \bibinfo{person}{Yangyang Fu}, \bibinfo{person}{Yixuan Wang}, \bibinfo{person}{Zhuoran Yang}, \bibinfo{person}{Zheng O'Neill}, \bibinfo{person}{Zhaoran Wang}, {and} \bibinfo{person}{Qi Zhu}.} \bibinfo{year}{2022}\natexlab{}.
\newblock \showarticletitle{Accelerate online reinforcement learning for building HVAC control with heterogeneous expert guidances}. In \bibinfo{booktitle}{\emph{Proceedings of the 9th ACM International Conference on Systems for Energy-Efficient Buildings, Cities, and Transportation}}. \bibinfo{pages}{89--98}.
\newblock


\bibitem[Yan et~al\mbox{.}(2018)]%
        {yan2018ridi}
\bibfield{author}{\bibinfo{person}{Hang Yan}, \bibinfo{person}{Qi Shan}, {and} \bibinfo{person}{Yasutaka Furukawa}.} \bibinfo{year}{2018}\natexlab{}.
\newblock \showarticletitle{RIDI: Robust IMU double integration}. In \bibinfo{booktitle}{\emph{Proceedings of the European Conference on Computer Vision (ECCV)}}. \bibinfo{pages}{621--636}.
\newblock


\bibitem[Yang et~al\mbox{.}(2022)]%
        {yang2022learning}
\bibfield{author}{\bibinfo{person}{Tsung-Yen Yang}, \bibinfo{person}{Justinian Rosca}, \bibinfo{person}{Karthik Narasimhan}, {and} \bibinfo{person}{Peter~J Ramadge}.} \bibinfo{year}{2022}\natexlab{}.
\newblock \showarticletitle{Learning physics constrained dynamics using autoencoders}.
\newblock \bibinfo{journal}{\emph{Advances in Neural Information Processing Systems}}  \bibinfo{volume}{35} (\bibinfo{year}{2022}), \bibinfo{pages}{17157--17172}.
\newblock


\bibitem[Yi et~al\mbox{.}(2022)]%
        {yi2022physical}
\bibfield{author}{\bibinfo{person}{Xinyu Yi}, \bibinfo{person}{Yuxiao Zhou}, \bibinfo{person}{Marc Habermann}, \bibinfo{person}{Soshi Shimada}, \bibinfo{person}{Vladislav Golyanik}, \bibinfo{person}{Christian Theobalt}, {and} \bibinfo{person}{Feng Xu}.} \bibinfo{year}{2022}\natexlab{}.
\newblock \showarticletitle{Physical inertial poser (pip): Physics-aware real-time human motion tracking from sparse inertial sensors}. In \bibinfo{booktitle}{\emph{Proceedings of the IEEE/CVF Conference on Computer Vision and Pattern Recognition}}. \bibinfo{pages}{13167--13178}.
\newblock


\bibitem[Zhang et~al\mbox{.}(2019)]%
        {zhang2019building}
\bibfield{author}{\bibinfo{person}{Chi Zhang}, \bibinfo{person}{Sanmukh~R Kuppannagari}, \bibinfo{person}{Rajgopal Kannan}, {and} \bibinfo{person}{Viktor~K Prasanna}.} \bibinfo{year}{2019}\natexlab{}.
\newblock \showarticletitle{Building HVAC scheduling using reinforcement learning via neural network based model approximation}. In \bibinfo{booktitle}{\emph{Proceedings of the 6th ACM international conference on systems for energy-efficient buildings, cities, and transportation}}. \bibinfo{pages}{287--296}.
\newblock


\bibitem[Zhang et~al\mbox{.}(2017)]%
        {zhang2017beyond}
\bibfield{author}{\bibinfo{person}{Kai Zhang}, \bibinfo{person}{Wangmeng Zuo}, \bibinfo{person}{Yunjin Chen}, \bibinfo{person}{Deyu Meng}, {and} \bibinfo{person}{Lei Zhang}.} \bibinfo{year}{2017}\natexlab{}.
\newblock \showarticletitle{Beyond a gaussian denoiser: Residual learning of deep cnn for image denoising}.
\newblock \bibinfo{journal}{\emph{IEEE transactions on image processing}} \bibinfo{volume}{26}, \bibinfo{number}{7} (\bibinfo{year}{2017}), \bibinfo{pages}{3142--3155}.
\newblock


\bibitem[Zhang et~al\mbox{.}(2023)]%
        {zhang2023modeling}
\bibfield{author}{\bibinfo{person}{Xiyuan Zhang}, \bibinfo{person}{Ranak~Roy Chowdhury}, \bibinfo{person}{Dezhi Hong}, \bibinfo{person}{Rajesh~K Gupta}, {and} \bibinfo{person}{Jingbo Shang}.} \bibinfo{year}{2023}\natexlab{}.
\newblock \showarticletitle{Modeling Label Semantics Improves Activity Recognition}.
\newblock \bibinfo{journal}{\emph{arXiv preprint arXiv:2301.03462}} (\bibinfo{year}{2023}).
\newblock


\bibitem[Zhu et~al\mbox{.}(2017)]%
        {zhu2017calibrating}
\bibfield{author}{\bibinfo{person}{Jincao Zhu}, \bibinfo{person}{Youngbin Im}, \bibinfo{person}{Shivakant Mishra}, {and} \bibinfo{person}{Sangtae Ha}.} \bibinfo{year}{2017}\natexlab{}.
\newblock \showarticletitle{Calibrating time-variant, device-specific phase noise for COTS WiFi devices}. In \bibinfo{booktitle}{\emph{Proceedings of the 15th ACM Conference on Embedded Network Sensor Systems}}. \bibinfo{pages}{1--12}.
\newblock


\bibitem[Zhussip et~al\mbox{.}(2019)]%
        {zhussip2019extending}
\bibfield{author}{\bibinfo{person}{Magauiya Zhussip}, \bibinfo{person}{Shakarim Soltanayev}, {and} \bibinfo{person}{Se~Young Chun}.} \bibinfo{year}{2019}\natexlab{}.
\newblock \showarticletitle{Extending Stein's unbiased risk estimator to train deep denoisers with correlated pairs of noisy images}.
\newblock \bibinfo{journal}{\emph{Advances in neural information processing systems}}  \bibinfo{volume}{32} (\bibinfo{year}{2019}).
\newblock


\end{thebibliography}

\end{document}